# A method for comparing multiple imputation techniques: a case study on the U.S. National COVID Cohort Collaborative

Elena Casiraghi[1,2,*], Rachel Wong[3,*], Margaret Hall[3], Ben Coleman[4,5], Marco Notaro[1,2], Michael D. Evans[6], Jena S. Tronieri[7], Hannah Blau[4], Bryan Laraway[8], Tiffany J. Callahan[8], Lauren E. Chan[9], Carolyn T. Bramante[10], John B. Buse[11,12], Richard A. Moffitt[3], Til Stürmer[13], Steven G. Johnson[14], Yu Raymond Shao[15,16], Justin Reese[17], Peter N. Robinson[4,5], Alberto Paccanaro[18,19], Giorgio Valentini[1,2], Jared D. Huling[20,**] and Kenneth J. Wilkins[21,**] on behalf of the N3C Consortium

[1] AnacletoLab, Department of Computer Science "Giovanni degli Antoni", Università degli Studi di Milano, Milan, ITALY
[2] CINI, Infolife National Laboratory, Roma, ITALY
[3] Department of Biomedical Informatics, Stony Brook University, Stony Brook, NY, USA
[4] The Jackson Laboratory for Genomic Medicine, Farmington, USA
[5] Institute for Systems Genomics, University of Connecticut, Farmington, CT, USA
[6] Biostatistical Design and Analysis Center, Clinical and Translational Science Institute, University of Minnesota, Minneapolis, MN, USA
[7] Department of Psychiatry, Perelman School of Medicine at the University of Pennsylvania, Philadelphia, PA, USA
[8] University of Colorado, Anschutz Medical Campus, Aurora, CO, USA
[9] College of Public Health and Human Sciences, Oregon State University, Corvallis, USA
[10] Division of General Internal Medicine, University of Minnesota, Minneapolis, MN, USA
[11] NC Translational and Clinical Sciences Institute, University of North Carolina at Chapel Hill, Chapel Hill, NC, USA
[12] Division of Endocrinology, Department of Medicine, University of North Carolina School of Medicine, USA
[13] Department of Epidemiology, Gillings School of Global Public Health, University of North Carolina at Chapel Hill, Chapel Hill, NC, USA
[14] Institute for Health Informatics, University of Minnesota, Minneapolis, MN, USA
[15] Harvard-MIT Division of Health Sciences and Technology (HST), 260 Longwood Ave, Boston, USA
[16] Department of Radiation Oncology, UT Southwestern Medical Center, Dallas, USA
[17] Environmental Genomics and Systems Biology Division, Lawrence Berkeley National Laboratory, Berkeley, CA, USA
[18] School of Applied Mathematics (EMAp), Fundação Getúlio Vargas, Rio de Janeiro, BRAZIL
[19] Department of Computer Science, Royal Holloway, University of London, Egham, UK
[20] Division of Biostatistics, School of Public Health, University of Minnesota, Minneapolis, MN, USA
[21] Biostatistics Program, Office of the Director, National Institute of Diabetes and Digestive and Kidney Diseases, National Institutes of Health, Bethesda, MD, USA

N3C Consortium: Tell Bennet, Christopher Chute, Peter DeWitt, Kenneth Gersing, Andrew Girvin, Melissa Haendel, Jeremy Harper, Janos Hajagos, Stephanie Hong, Emily Pfaff, Jane Reusch, Corneliu Antoniescu, Kimberly Robaski
Corresponding Authors (*): Elena Casiraghi, Rachel Wong
Co-senior Authorship (**): Jared D Huling and Kenneth J. Wilkins equally contributed to the work

**Abstract**

Healthcare datasets obtained from Electronic Health Records have proven to be extremely useful to assess associations between patients' predictors and outcomes of interest. However, these datasets often suffer from missing values in a high proportion of cases and the simple removal of these cases may introduce severe bias. For these reasons, several multiple imputation algorithms have been proposed to attempt to recover the missing information. Each algorithm presents strengths and weaknesses, and there is currently no consensus on which multiple imputation algorithms works best in a given scenario. Furthermore, the selection of each algorithm parameters and data-related modelling choices are also both crucial and challenging.

In this paper, *we propose a novel framework to numerically evaluate strategies for handling missing data in the context of statistical analysis*, with a particular focus on multiple imputation techniques. We demonstrate the feasibility of our approach on a large cohort of type-2 diabetes patients provided by the National COVID Cohort Collaborative (N3C) Enclave, where we explored the influence of various patient characteristics on outcomes related to COVID-19. Our analysis included classic multiple imputation techniques as well as simple complete-case Inverse Probability Weighted models. The experiments presented here show that our approach could effectively highlight the most valid and performant missing-data handling strategy for our case study. Moreover, our methodology allowed us to gain an understanding of the behavior of the different models and of how it changed as we modified their parameters. Our method is general and can be applied to different research fields and on datasets containing heterogeneous types.

**Introduction**

While electronic health records (EHRs) are a rich data source for biomedical research, these systems are not implemented uniformly across healthcare settings and significant data may be missing due to healthcare fragmentation and lack of interoperability between siloed EHRs [1][2].

Removal of cases with missing data may introduce severe bias in the subsequent analysis [3] and therefore imputing the missing information prior to conducting statistical analysis is often performed with the goal of reducing bias.

Imputation of missing data has been debated since the 1980s, when Rubin's seminal work [4] presented Multiple Imputation (MI) as an imputation strategy for statistical analysis. Based on Bayesian theory-motivated underpinnings [5][6], MI allows the natural variation in the data to be emulated in addition to accounting for uncertainty due to the missing values in the subsequent inferences. In practice, the objective of MI is to construct valid inference for the estimated quantity of interest [7] rather than being able to reconstruct/predicting the true missing values with greatest accuracy, which is the typical aim of imputation models applied in machine-learning contexts where the focus is on predictive analysis [8].

Since the introduction of MI, several MI algorithms have been proposed and successfully deployed in many different domains to avoid information loss before the application of standard statistical methods for causal inference [9][10] as well as machine learning techniques for predictive modeling [11][12].

However, there is no consensus on which MI algorithm works best under different scenarios. Aside from the choice of proceeding with an MI strategy (see Section Literature work), the choice of the specific imputation algorithm and of its input parameter settings, as well as modeling decisions such as the way datasets with heterogeneous types (categorical, numeric, binary) are handled are also both crucial and challenging. The appropriateness of including outcome variables in the imputation model also remains difficult to determine. For example, in a prior predictive study [11] the target (outcome) variable was omitted during imputation with the aim of avoiding bias in imputation results for variables highly correlated with the outcome. However, other works [13][14] recommend the inclusion of the outcome variables while estimating the missing values to control for confounders and obtain more reliable estimates.

In this paper, we consider statistical inference problems in the medical/clinical context and we focus on situations where (potentially adjusted) associations between patient characteristics and an outcome of interest needs to be inferred. In this context, we propose a method for evaluating and comparing several MI techniques, with the aim of choosing the most valid and performant approach for computing inferences in retrospective clinical studies. While we focus on the evaluation of MI algorithms, the method is general enough to be applied to any missing-data handling strategy.

To show the effectiveness and the practicability of the evaluation approach, we used as a case study a cohort of diabetic patients (type-2 diabetes) infected with COVID-19 provided by the National COVID Cohort Collaborative (N3C) Enclave (see Material and Methods). The same patient-cohort was previously filtered to remove cases with missing values and the obtained subcohort was analyzed to assess associations between hospital events (hospitalization, invasive mechanical ventilation, and death) and crucial diabetic patients' descriptors. Results of this analysis are reported in [15]. By limiting the analysis to complete cases, Wong et al. lost the 42% of cases therefore reducing the power of the estimator (subsection Case study: associations between diabetic-patients descriptors and COVID-19 hospitalization events). In literature, strategies such as Inverse Probability Weighting (IPW, [16][17], subsection Literature work) have been proposed to recover from the limits of complete-case analysis by including a "missingness model" into the overall analysis. However, frequently-applied implementations of these strategies, i.e., without augmentation [16], may prove ineffective when a high number of cases contain missing values. Imputation of missing data instead allows all cases to be used for computing potentially more reliable statistical estimates.

To guide the choice of the MI model and of its specification, we used our MI evaluation method to choose among the MI methods available from the N3C Platform.

Further, given its generality, we could apply the evaluation method to also assess the comparative performance of broader families of IPW models, comparing them to MI algorithms. In our case study, results confirmed the performance-based preference of multiple imputation over IPW models in this instance by comparing it with complete case analysis; such conclusions may not necessarily apply to augmented IPW that make use of all available data (including incomplete cases; for more on these so-called 'doubly-robust' IPW methods, see [17]) . To obtain actionable results, we finally used the most valid and performant (MI imputation) algorithms to compute odds ratios (and confidence intervals) describing associations between patient predictors and hospital events.

Of note, in [18] we applied our evaluation method to choose the most valid MI strategy in a context where a treatment effect must be estimated while adjusting for other, potentially confounding, variables (subsection Generalizability of the evaluation method to different scenarios, [18]). This is another practical example showing that our evaluation approach can be applied on a broad range of heterogeneous (clinical) datasets to compare different strategies and methods to handle missing data while performing statistical analysis.

The paper is organized as follows. In the Background section we first describe the case study we used to show the feasibility of the evaluation method (subsection Case study: associations between diabetic-patients descriptors and COVID-19 hospitalization events), and we next detail the MI strategy and its base theories (subsection Multiple Imputation), and a brief literature review (subsection Literature work). Next, section Evaluation method details our evaluation framework. The following section Experimental material and methods firstly reports the data source used for our experiments and implementation details (subsection Data source and implementation details), and then details the algorithms we evaluated on our case study and their experimental settings (subsection Experimented algorithms and settings); we conclude with Results, Discussion and Conclusions, and Highlights.

**Aim:** To propose an evaluation framework for comparing and contrasting different approaches for handling heterogeneous data missingness in the context of statistical analysis on real-world dataset.

**Statement of Significance**

**Problem:** Missing data is a problem affecting many research contexts. Imputation of missing data has been debated since the 1980s, when Rubin's seminal work presented multiple imputation as a key imputation strategy, given its ability to emulate the natural variation in data.

**What is Already Known:** Considering that multiple imputation strategies have produced promising results in many fields, numerous biomedical/clinical research works applied them to analyze patient data extracted from (electronic) health records. However, there are no established rules of thumb for choosing an effective multiple imputation method and any corresponding parameters for each such method.

**What this paper adds:** We propose an evaluation framework for comparing the validity and performance of multiple imputation algorithms in the context of retrospective clinical studies that assess potential associations between patient predictors and outcomes of interest. As a case study, we used the proposed evaluation method to compare different versions of mostly used multiple imputation methods over a cohort of diabetic patients (type-2 diabetes) provided by the National COVID Cohort Collaborative (N3C) Enclave. Besides the clinical usefulness of the obtained results, for each algorithm we evaluated various specifications, such as the inclusion/exclusion of the outcome variables in the imputation model, and the way categorical data is treated, among others. Beside testing the usefulness of the evaluation framework, the results allowed us to gain a better understanding of the behavior of the algorithms we compared. The generality of the evaluation approach allows it to be applied for assessing any (multiple) imputation procedure, including recent deep learning techniques that are gaining interest in several fields, as well as any strategy and algorithm for handling missing data in the context of statistical analysis, beyond those designed for data imputation.

# Background

*Case study: associations between diabetic-patients descriptors and COVID-19 hospitalization events*

We apply our methods to a previously published cases study [15] on patients with type 2 diabetes mellitus with data from the National COVID Cohort Collaborative (N3C). We used two logistic regression (LR) models and one Cox-Survival (CS) model to evaluate the association between glycemic control measured by Hba1c[1] and outcomes of acute COVID-19 infection, including mortality (hazards computed by a CS model), mechanical ventilation (odds computed by an LR model), and hospitalization (odds computed by an LR model). The study aimed at understanding the role of patient factors such as body mass index (BMI), race, and ethnicity on COVID-19 outcomes [20][21][22][23]. In particular, before running the LR and CS estimators, BMI was grouped according to the World Health Organization classification [24][25][2] that categorizes adults over 20 years of age as underweight (BMI < 18.5kg/m2), normal weight (18.5 ≤BMI < 25kg/m2), overweight ( 25 ≤ BMI < 30kg/m2), class I obesity (30 ≤ BMI < 35kg/m2), class II obesity (35 ≤ BMI < 40kg/m2), and class III obesity (BMI ≥ 40kg/m2) and the grouped variable was one-hot-encoded, so that the following estimators could evidence non-linear relationships between BMI and any of the three outcomes. Grouping and one-hot-encoding was also applied for the other numeric predictor variables (Hba1c and age). Table 1 reports details about the complete list of predictors, their type, the grouping of numeric variables, and the distribution of cases across all the predictors. Note that categorical predictors were also one-hot-encoded (Race[3], Ethnicity[4], Gender[5]) before the LR and CS analysis to explicitly investigate the influence of the different categories.

This study had several limitations. At first, it was only conducted using complete cases for whom data on height and weight were present to calculate BMI. In particular, 16,507/56,123 (29.4%) of patients were excluded due to missing BMI. In addition to BMI, race, and ethnicity information was also missing in the N3C data for a significant proportion of our cohort. To avoid the need to remove cases with missing ethnicity data, the authors introduced "race missing" and "ethnicity missing" as two additional categories that represented uncertain information and that were one-hot-encoded as the other race and ethnicity categories [6].

---

[1] HbA1c represents the integrated glucose concentration over the preceding 8–12 weeks [19].

[2] BMI is defined as a person's weight in kilograms divided by the square of their height in meters [24][25].

[3] In the N3C platform, the "Race" predictor reports whether the patient has race White, race Black or African American, Asiatic, he/she is Native Hawaiian or Other Pacific Islander, or has Other mixed race. In the cohort used by Wong et al. [15] no Native Hawaiian or Other Pacific Islander cases were found.

[4] In the N3C platform, the "Ethnicity" predictor reports whether the patient is Hispanic or Latino or Not Hispanic or Latino.

[5] In the N3C platform, the "Gender" predictor reports whether the patient is a "female", "male", or "other". The cohort used in [15] contained no cases with gender "other".

In particular 8643/56,123 (15.4%) of patients had missing data on race, 6,491/56,123 (11.6%) had missing data on ethnicity. This accounted for a total of 23,594/56,123 (42%) of samples containing missing or uncertain information. (Figure 1 shows details about the missing data pattern and the number of missing values per variable.)

In Wong's et al. cohort [15] the predictors were assumed to be Missing at Random (MAR), as suggested by Little's test [27], whose p-value (p<0.0001) allowed us to reject the null hypothesis of Missing Completely at Random (MCAR) missingness. Therefore, the listwise-deletion performed in the original analysis not only reduced the sample size and the statistical power of the estimator, but may have introduced bias in the computed inferences. Therefore, we chose to repeat the statical analysis described in [15], after a previous step where we imputed missing data in BMI[7], Ethnicity, and Race predictors.

*Multiple Imputation*

In the remainder of this paper, given a complete dataset $X \in R^{Nxd}$ (containing $N$ points represented by $d$ fully observed predictors) the statistical estimates (log odds and log hazard), their variance, standard error, and confidence interval will be referred to as $q_i$, $var_i$, $se_i$, and, $ci_i$, where the subscript $i \in \{1, \ldots, d\}$ will index the predictor variable. The notation used throughout the paper is summarized in Table 2.

When data contains missing values, the data may be *Missing Completely At Random* (MCAR), *Missing At Random* (MAR), or *Missing Not At Random* (MNAR) [28][29][30][31]. When the data are MCAR the missing observations are a (completely) random subset of all observations; in other words, the probability of being missing is uniform across all cases or, simply said, there is no relationship between the missing values and any other values, observed or missing. This implies that the missing and observed data values will have similar distributions. Consequently, apart from the obvious loss of information, the deletion of cases with missing values (generally referred to as "listwise deletion" or complete case analysis) may be a viable choice if the number of fully observed cases is sufficient to obtain reliable estimates.

---

[6] As this study was conducted within N3C there are tacit data curation features that provide a rationale for this: 1) NIH/NCATS data governance had an agreement with American Indian / Alaska Native sovereign tribal nations (through Summer 2022; see https://ncats.nih.gov/n3c/about/tribal-consultation) to deterministically impute 'Unknown' for their participants (to mitigate re-identification risk given concurrent availability of ZIP codes' first 3 digits), and 2) N3C consortium research indicates an increased risk for people of color to have incomplete mappings of race and/or ethnicity to an unambiguous harmonized OMOP set of fields, when other populations studies have demonstrated that these same subgroups-within-incompletely-mapped sites are at likely disparate risk of COVID-19 sequelae such as the outcomes studied [26].

[7] BMI is a dependent variable, with square dependency from height. To limit the effect of the square dependency, the logarithm of BMI is imputed and the resulting values are squared to revert to the original scale.

MAR data instead results in systematic differences between the missing and observed values, but these can be entirely explained by observed values in other variables (thus MCAR can be viewed as a more restrictive special case of MAR). In this case the probability of being missing is the same only within groups defined by the observed data (i.e., cases with missing values occur 'at random' within latent groups determined by observed variables), which means that there are relationships between missing and observed values, and these relationships may be exploited by proper data imputation techniques to compute valid inferences for the missing data.

In contrast to MCAR data, for MAR data the removal of cases with missing values can affect statistical power [3] and can introduce severe bias [32][33]. Indeed, for MAR data Little and Rubin [32] showed that the bias in the estimated mean increases with the difference between means of the observed and missing cases, and with the proportion of the missing data. Schafer and Graham [33] reported simulation studies where the removal of cases with missing values introduces bias under both MAR and MNAR missingness.

MNAR data is present when the data is neither MCAR nor MAR, wherein missingness depends on unobserved data (thus MAR can be viewed as restrictive special cases of MNAR where dependence on unobserved data no longer holds). In this case, missingness is not at random, and it must be explicitly modeled to avoid some bias in the subsequent inferences [34][35]. As MNAR is unverifiable and, in fact, non-identifiable from observed data [35], for any specific model of a MNAR mechanism to be adopted within an analysis, it must be postulated using domain-expert-driven assumptions. Given this context-specific aspect of MNAR modeling, we consider it outside the scope of this paper (aimed at proposing a generic evaluation framework for methods accommodating MAR missingness in EHR-based data) to posit specific MNAR mechanisms, as they are indistinguishable from specific MAR mechanisms given a set of observed data; sensitivity analysis frameworks or other methods necessarily specific to a particular research question, and beyond the scope of this work, should be used to assess MI or IPW techniques under a particular MNAR assumption.

When an (univariate or multivariate) MI strategy is chosen for imputing the missing values prior to conducting the analysis, the following three steps are consecutively applied (sketched in Figure 2).

(1) An (univariate or multivariate) imputation algorithm containing some randomness is used to impute the dataset a number $m$ of times, therefore obtaining a set of $m$ imputed sets, $\widehat{X^{(1)}}, \ldots, \widehat{X^{(j)}}, \ldots, \widehat{X^{(m)}}$, where the superscript $j \in \{1, \ldots, m\}$ will be used in the remaining part of this work to index the imputation number.

(2) Each of the $m$ imputed datasets is then individually analyzed to obtain a vector $\widehat{Q^{(j)}} = \left[\widehat{q_i^{(j)}}\right]$ of estimates for each predictor variable (indexed by the subscript $i \in \{1, \ldots, d\}$). Together with the estimates, the vector of variances of the estimates, $\widehat{VAR^{(j)}} = \left[\widehat{var_i^{(j)}}\right]$, their standard errors, $\widehat{SE^{(j)}} = \left[\widehat{se_i^{(j)}}\right]$, and confidence intervals $\widehat{CI^{(j)}} = \left[\widehat{ci_i^{(j)}}\right]$ are estimated.

(3) the $m$ estimates are then pooled by Rubin's rule [4] to obtain the final pooled inference as the mean of the estimates across all the imputations: $\hat{Q} = \frac{1}{m}\sum_{j=1}^{m} \widehat{Q^{(j)}}$ and its total variance, $\hat{T} = \widehat{W} + \left(1 + \frac{1}{m}\right)\hat{B}$, where $\widehat{W} = \frac{1}{m}\sum_{j=1}^{m} \widehat{VAR^{(j)}} \approx W_\infty$ is the estimate of the (true) within imputation variance (that would be obtained when $m \to \infty$) and $\hat{B} = \frac{1}{m-1}\sum_{j=1}^{m}\left(\widehat{Q^{(j)}} - \hat{Q}\right)^2 \approx B_\infty$ is the estimate of the (true) between imputation variance (when $m \to \infty$). $\hat{T} \approx T_\infty$ is an estimate of the true variance obtained when $m \to \infty$.

While easy to define in principle, the specification of a multiple imputation pipeline is not easy, since several open issues remain to be clarified. First, there does not exist a clear and well-defined theory that allows the optimal number $m$ of multiple imputations to be chosen. Indeed, several researchers [36][37] have supported Rubin's empirical results [4] according to which 3 to 10 imputations usually suffice for obtaining reliable estimates. However, more recent research [38][39] has reported experimental results showing that the number of multiple imputations should be set to larger values (e.g. $m \geq 20$), which is now computationally more feasible than it was several decades ago.

In our settings, $m$ was chosen in order to maximize the efficiency of the multiple imputation estimator (see Appendix A). This was obtained by applying Von Hippel's [40] rule of thumb, according to which a number of imputations comparable to the percentage of cases that are incomplete would allow maximizing the efficiency of the estimator. When amputating the data to reproduce the presumed-MAR patterns in our dataset, this criterion required setting $m = 42$; however, since the definition of $m$ is controversial and no well-accepted rule has been defined, we also performed experiments with the value $m = 5$ suggested by Rubin and set as default by many packages (section Results). This allowed the stability of the computed estimates to be assessed with respect to the value of $m$.

Secondly, beside the arduous choice of the imputation algorithm, its application settings are also both crucial and challenging. This choice depends on the data structure, the data-generating mechanism, the inferential model, and the scientific question at hand.

*Literature work*

Statistical analysis of incomplete (missing) data is gaining a lot of interest in the research community. To this aim, classic approaches such as simple complete-case Inverse Probability Weighting (IPW, [16]) limit the statistical analyses to the subset of complete cases weighted by their inverse probability of containing missing values. Since this probability is often unknown, it is generally estimated by using a logistic regression model that is fitted on the complete predictors and with outcome given by an indicator of each case containing at least one missing value. Though effective in several contexts, when a complex missingness pattern is present in the data and many predictors contain missing values, or when many cases are incomplete, IPW models tend to have a significant power loss due to the high number of cases being dropped; beyond the scope of this paper, the degree of power loss when using augmented IPW approaches leveraging all available data, e.g., [41], would warrant separate lines of research.

In these contexts, (multiple) data imputation strategies have often proven their effectiveness. In particular, MI algorithms can broadly be classified into three categories:
1. parametric multivariate MI imputation techniques exploiting a Joint Modeling (JM) approach [7][37];
2. univariate imputation methods exploiting a Fully Conditional Specification (FCS) strategy [28];

3. machine learning-based (e.g., missForest [42]) or deep-learning based MI strategies (e.g., MI via autoencoder models [43] – e.g. MIDA [44], or stacked deep denoising-autoencoders [45], or Generative Adversarial Networks [46] – e.g. GAIN [47] or MisGAN [48]).

Multivariate MI techniques exploiting a JM imputation strategy assume a joint distribution for all variables in the data and generate imputations for values in all variables by drawing from the implied conditional (predictive) distributions of the variables with missing values [37]. The multivariate JM strategy adheres to Rubin's theoretical foundations [4] and its empirical computational time costs are significantly lower than those required by univariate FCS imputation algorithms (see below); however, it is often challenging to specify a joint underlying distributional model, and this particularly happens when dealing with high-dimensional datasets and/or datasets characterized by mixed variable types (including binary and categorical types). For these reason, some of the most popular algorithms exploiting a multivariate-JM strategy (e.g., "norm", the classic MI multivariate imputation function implemented in R language [37], PROC MI [49], and Amelia [50] - subsection <u>Experimented algorithms and settings</u>), simplify the problem by assuming a multivariate Gaussian distribution as the underlying distribution, and apply the same strategy also for categorical variables, after converting then to numeric (integer) variables. Besides implicitly imposing an ordering between categories this can lead to bias, as documented by [51]. To avoid any simplifying assumptions about the joint distribution, flexible and nonparametric techniques have been proposed [7] that obtain effective imputation results by modeling the joint distribution through advanced Bayesian techniques. However, they incur high computational costs, hampering their practical applicability on high-dimensional, complex datasets such as those recently available from medical EHR studies.

FCS approaches exploit a univariate imputation approach where an univariate conditional distribution (generally the normal distribution) is defined for each variable with missing values given all the other variables. This allows designing an iterative procedure where missing values are imputed variable-by-variable, akin to a Gibbs sampler. The most representative among MI algorithms using the FCS strategy is Mice [52] (subsection <u>Experimented algorithms and settings</u>); it initially imputes the missing data in each variable by using a simple hot-deck-imputation technique (the mean/mode of observed values), and then imputes each incomplete variable by a separate model that exploits the values precedingly imputed from the other variables to "chain" all the univariate imputations. By default, Mice uses predictive mean matching (pmm, [28]) for imputing missing values in numeric data, logistic regression and polytomous logistic regression for binary data or categorical data. However, its version using classification and regression trees (CART, [53][54]) has also achieved promising results [55], as CART and regression trees are more-flexible estimation procedures. Other flexible machine-learning and deep-learning-based imputation techniques show promise as well.

MissForest [42] is a notable such procedure. It imputes missing values by applying an univariate FCS strategy, where variables with missing values are imputed by using RFs [56] for either regression (integer- or real-valued variables) or classification (binary or categorical variables). MissForest was presented as an imputation method to be applied for predictive modeling, where a unique imputation of missing data is generally produced before training any subsequent classifier on the imputed data. However, an MI version of MissForest was proposed in missRanger, where a final refinement step is added that applies pmm[8] to both avoid outliers and recover the natural data variability (subsection Experimented algorithms and settings).

Given the success of deep-learning techniques in several fields, several authors have designed flexible deep, neural-network-based imputation models that have shown promise even in the presence of complex data [8]. In particular, two recent advances in the context of deep-neural networks are particularly suited for the task of (multiple) data imputation: denoising autoencoders (for MI) [43][44][45] and Generative Adversarial Networks (GANs) [46][47][48].

In particular, autoencoders are unsupervised neural networks that compute an informative lower-dimensional representation of the input data. They are generally characterized by an hourglass shaped architecture, composed of two modules: an encoder-module and a decoder-module connected that share a bottleneck layer. The encoder-module processes the input layer to produce a lower dimensional representation of the input data in the so-called bottleneck layer; the decoder module processes the output of the bottleneck layer (the lower dimensional input representation) to obtain an output layer that best reconstructs the input data. After being trained by a loss function that measures the difference between the input and the output layers, the autoencoder can be used to process input samples to retrieve their lower dimensional representations in the bottleneck layer. In practice, an autoencoder is a neural network model trained to learn the identity function of the input data. Denoising autoencoders intentionally corrupt the input data (by randomly turning some of the input values to zero) in order to prevent the networks from learning the identity function, but rather a useful low-dimensional representation of the input data. Given a sample with missing values, denoising autoencoders are naturally suitable for producing MI data because they can simply be run several times by using different random initializations. A classic example of algorithm using denoising autoencoder for MI is MIDA [44][9].

---

[8] When a pmm model exploiting k donors is used to assign a continuous or categorical label to a test sample, $x_{testRd}$, the k training points (donors) that are the nearest to $x_{test}$ (according to a proper similarity metric) are selected and the label of a randomly chosen donor is assigned to $x_{test}$.

GANs [46][47][48] are other neural-network models generally used for generative modeling, that is to output new examples that plausibly could have been drawn from the original dataset.

GANs allow reformulating the generative model as a supervised learning problem with two sub-modules: a generator-module that is trained to generate new examples, and the discriminator-module that is trained to classify examples as either real (i.e., from the domain to be learnt) or fake (generated). The two models are trained together in an "adversarial" game, until the discriminator-module is fooled about half the time, meaning that the generator-module is generating plausible examples. GAIN [47] and MisGAN [48] are two recent examples of MI algorithms that utilize GANs. Given an input dataset with missing values, both of them first add noise and fill the missing values with some hot-deck imputation technique or constant values.

Next, GAIN [47] employs an imputer-generator module that is trained to produce plausible imputations of the missing data. The generator is trained adversarially with a discriminator that determines which entries in the completed data were actually observed and which were imputed.

MisGAN [48] instead uses a generator-module that is trained to generate both plausible imputation values and missingness masks that mark the values that have been imputed. This generator is adversarially trained with a discriminator-module that solely works on the masked-output of the generator to recognize a valid imputed version of the data. Both GAIN and MisGAN can be used to generate MIs by using several runs of the GAN model with varying initial noise and/or imputations of the missing values.

Machine-learning and deep-learning based MI methods have three main advantages over traditional multivariate-JM and univariate-FCS MI models. First, they are more flexible and do not need any underlying data distribution to be specified. Second, they are naturally designed to deal with mixed data-types. Third, they can uncover more-complex, nonlinear relationships between variables and are able to exploit them to improve the validity of the computed imputations. Deep-learning based MI methods are further characterized by their documented ability to impute complex, high-dimensional data. Moreover, once trained, the computational time of deep-learning based models is much lower than that required by univariate-FCS algorithms (e.g., MICE) and univariate machine-learning based algorithms (e.g. missForest).

---

[9] MIDA is available as an R package: https://cran.r-project.org/web/packages/rMIDAS/rMIDAS.pdf

These advantages are however counterbalanced by crucial points, often hampering the practical application of deep-learning based MI techniques; indeed, the hyperparameter tuning of deep-learning based models is difficult and crucial, and slightly different model architectures can result in dramatically different results. Unfortunately, few details are provided about hyperparameter tuning and architectural choices, and their consequences for the performance of imputation methods. For this reason, their choice is generally limited to predictive modeling contexts, where the choice of the best architecture and hyperparameter values may be simply guided by the prediction performance.

Further, while considering recently proposed non-MI and MI methods using deep-learning models to handle simulated MNAR data [57][58], one encounters a lack of transparency in how deep architectures encode the missingness mechanism assumptions. Indeed, the proposed models tacitly adopt the (unidentifiable) MAR/MNAR assumptions, akin to failing to accommodate how two- and higher-way interactions among discrete sets of patient-level predictors might impact outcomes and analytic features deemed crucial by clinicians in most settings.

This inability to accommodate clinically-valid joint variable distributions with explainable distinctions between MAR and MNAR, together with our implementation challenges above, highlight that the proper setting of deep-learning models in the context of MI is still lacking a grounded theoretical basis, whose definition would require deep research investigations.

**Evaluation method**

In statistical inference contexts, the goal of MI is to obtain statistically valid inferences from incomplete data. In other words, given a statistical model of interest (e.g., an LR estimator or a CS model), an imputation algorithm should ultimately allow the user to obtain estimates as similar as possible to those that the statistical model would provide if the data were complete.

Unfortunately, there is no rule of thumb for choosing an imputation model based on the problem at hand, the amount of missingness, or the missingness pattern.

However, following the guidelines in Van Buuren's seminal work [28], when the number of complete cases has a reasonable cardinality, an MI algorithm may be evaluated by comparing the inferences (i.e. statistical estimates) obtained on the dataset containing fully observed data (complete dataset obtained by listwise deletion) to those computed by pooling all the MI estimates obtained on an *amputated* version of the complete dataset, where *amputation* refers to the process that synthetically generates missing values in a dataset [59]. Our framework leverages Van Buuren's guidelines and proposes a set of evaluation measures that are pooled across multiply amputated datasets.

In practice, given a statistical estimator of interest, we propose to numerically evaluate a specific MI algorithm by applying the steps sketched in Figure 3:

(step 1) starting from the original dataset with missing values, obtain a *complete dataset* by listwise deletion and apply the statistical estimator of interest to compute gold standard estimates;

(step 2) produce *A* amputated versions of the complete dataset by reproducing the same (MCAR or MAR) missingness observed in the original dataset (for details about the best value for parameter *A* see Section Results, while a discussion about proper approaches for reproducing MCAR or MAR missingness is reported below);

(step 3) process each amputated dataset by applying the MI estimation pipeline detailed in Section Multiple Imputation (sketched in Figure 2) to obtain a vector of MI estimates (one estimate per predictor variable in the dataset);

(step 4) average each MI estimate across all the amputations to obtain a vector approximating the expected value of the MI estimate for all the predictor variables in the dataset. Finally compare the gold standard estimates (obtained on the complete dataset) to the expected values of the MI estimates by using the numeric evaluation measures detailed below.

Before detailing the evaluation method, an important note about the above step-2 is due, which regards the simulation of the (MAR or MCAR) missingness pattern in the available dataset.
Reproduction of MCAR missingness in the complete dataset is simple and requires producing the same missingness proportions by sampling from uniform distributions.
On the other hand, data amputation to simulate a MAR mechanism is a difficult and challenging procedure and few works describing different methodologies for producing simulated MAR data are available in literature [59]. The crucial point is the estimation of the distribution describing the missingness in one variable conditioned to other variables. The literature review in this context highlights that the method proposed by Shouten et al. [59] is the most promising and reliable to emulate the MAR missingness characterizing a given dataset. Moreover, Shouten et al. support their results with extensive simulations showing the effectiveness of the produced amputations. Therefore, we produced MAR missingness by using the function "ampute", available from the MICE package R, which implements the Shouten's et al. techniques[10].

We must further note that it is possible for the subset of complete data to differ systematically from the full data that have "real" missingness, so that any amputation procedure would not be able to reproduce the exact missingness. However, this numerical approach of evaluating imputation methods still has merit as an evaluative technique, as it provides empirical evidence of the ability of imputation approaches to handle observable missingness patterns in a given dataset.

Hereafter, we report the details about the evaluation method and the evaluation metrics we are using to compare the gold standard estimates to the MI estimates. To aid comprehension, Figure 4 reports a more detailed overview of the steps we are applying and consider the more general situation when more than one statistical estimator is applied, where each estimator computes inferences related to a specific outcome of interest (as an example, in our use case we had three outcome of interest and respective statistical estimators). In particular, for each outcome variable (statistical estimator) the following steps are applied to evaluate an MI algorithm:

1) *Obtain gold standard statistical estimates (and their confidence intervals) on a complete dataset* [light green box on top of Figure 4]. To this aim, listwise deletion is applied to get the *complete dataset* $X \in R^{N \times d}$ (composed by $N$ complete cases and $d$ predictor variables). The complete dataset is then *normalized* to have predictors with the same scale, and the estimator of interest is applied to get a vector of statistical estimates for each predictor variable in the dataset $i \in \{1, \ldots, d\}$. Note that the dataset normalization step is not mandatory but it ensures obtaining estimates characterized by the same scale. This is a useful characteristics (as will be shown below).

---

[10] Further details about the "ampute" function are reported at:
https://rianneschouten.github.io/mice_ampute/vignette/ampute.html#Introduction_to_mice::ampute
https://www.gerkovink.com/Amputation_with_Ampute/Vignette/ampute.html#:~:text=The%20function%20ampute%20works%20by,a%20certain%20missing%20data%20

Hereafter, the vector of gold standard statistical estimates will be referred to as $Q = [q_i]$ and the corresponding vector of confidence intervals will be $CI = [q_i^{min}, q_i^{max}] = [CI_i]$.

2) *Obtain the expected values of the MI estimates* [light-blue box in Figure 4]. To this aim, the following steps are applied:

I. Compute $A$ amputations of the complete datasets (step I). Hereafter, we will refer to the $j$th amputated dataset as $\widehat{X^{(j)}}$ ($j = 1,\ldots,A$).

At this stage, the MI estimation pipeline (Section Multiple Imputation and Figure 2) is applied to each of the $A$ amputated datasets. More precisely, on the $a$-th ($a \in \{1,\ldots,A\}$) amputated dataset:

II. the MI algorithm under evaluation is applied to compute $m$ imputations

III. each imputed dataset is *normalized* as done on the complete dataset (light-green box in Figure 4) to obtain predictors with the same scale.

IV. the statistical estimator of interest is applied on each imputed-normalized dataset and Rubin's rule is used to pool all the estimates and obtain an imputation estimate. This allows to compute a vectors of estimates $\widehat{Q(a)} = [\widehat{q_i(a)}]$, the vector of standard errors, $\widehat{SE(a)} = [se_i(a)]$[11], and the vector of the 95% confidence interval, $CI(a) = \widehat{[ci_i(a)]}$ [12] ($i \in \{1,\ldots,d\}$) for the $a$-th amputated dataset. The vector is composed of the imputation estimates for each of the $d$ predictors in the dataset.

V. For each predictor variable, average all the estimates across the $A$ amputated datasets to obtain a vector approximating the expected value of the MI estimate $E[\hat{Q}]$. The vector $E[\hat{Q}]$ is computed as: $\underline{Q} = [q_i] \approx E[\hat{Q}] = \frac{1}{A}\sum_{a=1}^{A}\widehat{Q(a)}$

3) *Compare the gold standard estimates to the MI estimates* [yellow-box in Figure 4]: the inferences obtained on the complete data ($Q = [q_i]$ - step 1) are compared to those obtained on each of the amputated datasets $(\widehat{Q(a)} = [\widehat{q_i(a)}], i \in \{1,\ldots,d\})$ by computing the evaluation measures

---

[11] The standard error of the estimates is computed via the "pool" function provided by the Mice package, which exploits Rubin's rule [4] to compute both the total variance and the standard error of the estimate.

[12] The CI of the estimates is computed by using the "confint" function from stats package in R. The function uses a maximum likelihood estimator.

described below (some of which are listed in [23] by considering a unique amputated dataset). In particular, considering the vectors containing all the predictor estimates, we compute

- the **raw bias** vector $RB = [rb_i]$ $(i \in \{1,\ldots,d\}) = \underline{Q} - Q \approx E[\hat{Q}] - Q$, where $rb_i$ is the raw bias for the $i^{th}$ predictor variable, whose sign may be observed across all the predictor variables to understand whether the multiple imputation has the effect of globally underestimating or overestimating the true estimates. This information is complemented by the **estimate ratio (ER)**, $ER = [er_i]$ $(i \in \{1,\ldots,d\}) = \frac{\underline{Q}}{Q} \approx \frac{E[\hat{Q}]}{Q}$, being $er_i$ the estimate ratio for the $i^{th}$ predictor variable, and by the vector containing the **expected value of the Mean Squared Error (MSE)** of the estimate, where again $mse_i$ refers to the $i^{th}$ predictor: $MSE = [mse_i] = \frac{1}{A}\sum_{a=1}^{A}(\widehat{Q(a)} - Q)^2 \approx E[(\hat{Q} - Q)^2]$.

- the **coverage rate ($cr_i$)** for the $i^{th}$ predictor variable is the proportion over all the $A$ amputations of the confidence intervals $\widehat{ci_\iota(a)} = $ (for each $a \in \{1,\ldots,A\}$) that contain the true estimate $q_i$. The actual rate should be equal to or exceed the nominal rate (95%). If $cr_i$ falls below the nominal rate, the method is too optimistic, leading to higher rates of false positives. A too low $cr_i$ (e.g. below 90 percent for a nominal 95% interval) indicates low reliability. On the other hand, a too high $cr_i$ (e.g., 0.99 for a 95% confidence interval) may indicate that the confidence intervals of the pooled estimates $\widehat{ci_\iota(a)}$ are too wide, which means that the MI method could be inefficient. In this case, the analysis of average standard error of each pooled estimates, $\widehat{se_\iota(a)}$, and the ratio of the standard errors of the pooled and true estimates $ratio\ se_i = \frac{\widehat{se_\iota(a)}}{se_i(a)}$ may inform about the effective reliability of a high $cr_i$ value. In practice high values of $cr_i$ are consistent with standard errors $\widehat{se_\iota}$ of the MI estimate lower or comparable to the standard error of the true estimates. In other words, high values of $cr_i$ are valid when $ratio\ se_i \leq 1$. To obtain such result the number of imputations must be as high as to guarantee that the variance of the MI estimate is mainly dominated by the variance of the statistical estimator (essentially evaluated by the within imputation variance).

We note that, when considered by itself, each of the numeric measures described above provides limited information about the validity of the MI algorithm being evaluated. However, when considered together the evaluation measures we are proposing provide a full picture about the capability of each MI algorithm to provide estimates approximating the gold standard estimates computed on the complete set.

In any case, the proposed evaluation measures are general and can be used under different scenarios (subsection Generalizability of the evaluation method to different scenarios) where statistical estimators providing estimates, standard error of the estimate, and confidence intervals must be applied. These procedures and metrics have been applied to evaluate the validity of the estimates computed by IPW models and to compare them to those computed by MI algorithms (subsection Experimented algorithms and settings and section Results).

Note that, until the past decade, several methods amputated the complete data and then compared the application of different imputation algorithms by using the RMSE (or NRMSE) between the true values and the imputed values. However, literature works have cleared that such metrics may not give a full picture of the comparisons and sometimes can be even misleading and may lead to obscure, unreliable conclusions [60]. Moreover, when redesigned for assessing MI datasets, they are usually computed by considering the mean across all the imputations, which results in an opaque metric with an uncertain statistical meaning that ignores the uncertainty of imputations.

**Experimental material and methods**

*Data source and implementation details*

The dataset used in this study, has been provided by the N3C Enclave. The N3C receives, collates, and harmonizes EHR data from 72 sites across the US. With data from over 14 million patients with COVID-19 or matched controls, the N3C Platform (©2021, Denver, CO) provides one of the largest and most representative datasets for COVID-19 research in the US [61][62][63].

The rationale, design, infrastructure, and deployment of N3C, and the characterization of the adult [62] and pediatric [63] populations have been published previously. Continuously updated data are provided by health care systems to N3C and mapped to the OMOP common data model[13] for authorized research.

---

[13] https://ohdsi.github.io/TheBookOfOhdsi/

N3C data has been used in multiple studies to better understand the epidemiology of COVID-19 and the impact of the disease on health and healthcare delivery [15][64][65][66][67][68].

All the code for the analysis was implemented on the Palantir platform leveraging the Foundry operating system[14]. The platform enables groups of users to share code workbooks. Each code workbook is articulated into communicating nodes; each node can be written in SQL, Python/Pyspark, or R/RSpark code. The input and output of each node must be formatted in the form of a table (tabular dataframe) or a dataset (a collection of tables/dataframes).

For consistency, all IPW [13] and imputation algorithms we used (detailed in the following subsection Experimented algorithms and settings) are implemented in R packages, available from the CRAN repository. These included Amelia (version 1.7.6,[50][15]), Mice (version 3.8.0, [52][16]), and missRanger (version 2.1.3, [42][17]).

*Experimented algorithms and settings*

To show the feasibility and practicability of our evaluation method, we conducted a series of experiments where IPW models were compared to MI techniques for obtaining statistical estimates on the use-case presented in subsection Case study: associations between diabetic-patients descriptors and COVID-19 hospitalization events.

The IPW models varied for the method used to compute the probability of missing values and for the predictors used to estimate it. More precisely, for probability estimation we compared the usage of logistic regression models and random RFs [56]. For what regards the predictor variables that were used to estimate the missingness probability, we compared the inclusion/exclusion of the outcome variables in the prediction model. Moreover, in line with the LR models applied by Wong et al. [15], we also compared the usage of numeric variables (age and Hba1c) to the setting where numeric variables are binned and then one-hot-encoded.

---

[14] https://www.palantir.com/platforms/foundry/
[15] https://cran.r-project.org/web/packages/Amelia/Amelia.pdf
[16] https://cran.r-project.org/web/packages/mice/mice.pdf
[17] https://cran.r-project.org/web/packages/missRanger/missRanger.pdf

The MI algorithms were chosen among those that 1) obtained good performance as reported by literature studies and by preliminary experiments, 2) were freely available in three MI packages, 3) had a memory/time complexity supporting the computation on a large dataset within the N3C Palantir secure analytics platform, and, above all, 4) applied different strategies (FCS imputation, JM imputation, or machine-learning based imputation) and were based on different theoretical grounds and assumptions.

Briefly, two imputation algorithms, Amelia [50] and Mice [52] exploit, respectively, a multivariate-JM strategy and an univariate-FCS strategy where an underlying normal distribution is assumed; the third method (missRanger [42]) is a representative of more-flexible machine-learning-based imputation approaches. All the methods are described in Literature Work. In the following we describe their different specifications we have compared by using the evaluation method proposed in this paper (Section Evaluation method).

In its default settings, for each variable with missing values, *Mice* uses the observed part to fit either a predictive mean matching (pmm, for numeric variables), or a logistic regression (for binary variables), or a polytomous regression (for categorical variables) model and then predicts the missing part by using the fitted model.

Mice also provides the ability to exploit a Bayesian estimator for imputing numeric variables. To perform an exhaustive comparison, we therefore compared the performance of Mice with default settings (referred to as *Mice-default* in the following) with those of a Mice using univariate Bayesian estimators (hereafter denoted with *Mice-norm*) and run on a version of the dataset where all the categorical variables are one-hot-encoded to convert them to a numeric type.

This choice is coherent with the study from [15], where authors performed their analysis by 1) one-hot-encoding categorical variables (e.g. "Race"), and 2) binning continuous variables (BMI, age, and hba1c) and then one-hot-encoding the resulting binned variables. As aforementioned, using such representation in an imputation setting essentially results in a fuzzy imputation, where each imputation run is allowed to choose multiple categories for each sample. We thus aim to understand if one-hot-encoding both numeric and categorical variables could positively or negatively affect the obtained results. Experimenting this setting with *Mice-default* resulted in the application of a logistic regression model for each binned variable to be imputed. We use ***Mice-logreg*** to denote the Mice algorithm run on samples expressed by one-hot-encoded variables imputed via logistic regression models. For exhaustiveness of comparison we also used the application of Mice-norm under this setting.

Independent from the exploited univariate imputation models, all the Mice algorithms iterate their univariate imputations over all the variables with missing values by following a pre-specified variable imputation order (increasing or decreasing number of missing values), and then restart the iteration until a stopping criterion is met, or a maximum number of user-specified iterations is reached. The reason for the multiple iterations is that, from iteration 2 onwards, each model refines the previous imputations by exploiting the better quality data that has been previously imputed. In our experiments, due to the high sample cardinality, we allowed a maximum number of iterations equal to 21 and we tested the application of Mice when the univariate imputation order was given by the increasing and the decreasing number of missing values.

The missRanger algorithm (hereafter denoted ***missRanger***) is a fast R implementation of the missForest algorithm, which applies the same univariate, iterative imputation schema used by Mice, where the main difference is in the usage of the RF model for each univariate imputation. Note that, in between the consecutive variable imputations, missRanger allows using the pmm estimator (Section <u>Literature work</u>). In this way, for each imputed value in variable $v$, pmm finds the nearest $k$ predictions for the observed data in $v$, randomly chooses one of the $k$ nearest predictions, and then uses the corresponding observed value as the imputed value. The application of pmm firstly avoids imputing with values not present in the original data (e.g. a value less than zero in variables with non-negative valued variables); and secondly, it allows raising the variance in the resulting conditional distributions to a realistic level.

In our experiments, due to the high sample cardinality, we used 50 trees per RF, allowed a maximum number of iterations equal to 21, tested the application of missRanger by using the univariate imputation order given by the increasing and the decreasing number of missing values, and we also compared the behavior of the algorithm imputation results when predicting mean matching is avoided (***missRanger no-pmm***), or when it is applied with 3 or 5 donors (values suggested by the authors). Further, for allowing an exhaustive comparison to the setting where all the variables are binarized we also tested the application of missRanger under the scheme when categorical and binned numeric variables are one-hot-encoded.

The *Amelia* algorithm uses the Expectation Maximization algorithm presented in [50] to estimate the parameters underlying the distribution behind the complete observations, from which the imputed values are drawn. In case of categorical data, Amelia uses a one hot encoding strategy, which essentially reverts to fuzzy imputations for categorical variables. Similar to what was done for the other experimented algorithms, we also experimented after binning and one-hot-encoding numeric variables.

In Table 3 we detail the imputation algorithms used for the experiments and the settings we compared. By testing the three imputation algorithms (Mice, missRanger, and Amelia) and assessing the differences obtained when (1) considering/neglecting the outcome variables in the imputation model, (2) one-hot-encoding numeric and categorical variables or keeping their natural type, (3) varying the univariate imputation order (for the missRanger and the Mice methods), (4) varying the number of pmm donors in missRanger, we obtained 44 different MI specifications.

## Results

When evaluating the various missing-data handling algorithms (subsection <u>Experimented algorithms and settings</u>), we considered the same patients' cohort detailed in [15], we obtained a complete dataset by listwise deletion, and we run our evaluation pipeline by using $A = 25$ amputated datasets, where we simulated MAR missingness with similar missingness patterns.

In our experiments we empirically set the value $A = 25$. This value was chosen as a tradeoff between the computational memory/time complexity supported by the N3C Platform and the stability of the obtained results. To choose it, we started by a low number of amputated datasets ($A = 5$) and we increased the number until we noted no appreciable changes of the computed measures. Obviously, having high computational power, a higher number of amputated datasets would be suggested to guarantee robustness of the obtained results.

For the evaluation, we defined a statistical estimation pipeline that reproduces the analysis in [15].

More precisely, a first step of variable binarization[18] was applied to have comparable scales across predictors (both in the imputed and in the complete dataset) and the normalized dataset was then used to run two LR models and one CS model to understand the influence of the available predictors on the hospitalization event, the invasive ventilation - i.e. mechanical ventilation - event, and the patients' survival. These analyses constituted the scientific analyses of interest in the motivating study and thus served as the basis for evaluating the different missing-data handling approaches. In particular, the parameters estimated by these three models (i.e. the log odds ratios) served as the targets of estimation. We treated the estimates of these parameters on the complete dataset (and their associated standard errors and confidence intervals – top left forestplots in Figure 5, Figure 6, Figure 7) as the gold standard and compared our estimates of these parameters using the amputated data to these gold standard estimates.

The evaluation measures described above were computed for all the considered IPW algorithms and MI algorithms under their different specifications (subsection <u>Experimented algorithms and settings</u>). As a result, for each algorithm we obtained evaluation measures for $d = 38$ binarized predictor variables and $O = 3$ estimates.

In tabular Figure 8 we provide a visual comparison of the computed evaluation measures, where the number of imputations ($m = 42$) was chosen according to Von Hippel's rule of thumb [40] (Appendix A). To assess the significance of the comparison between different (IPW and MI) algorithms, we applied the two-sided Wilcoxon signed-rank test at the 95% confidence (p-value < 0.05). For the sake of exhaustiveness, the win-tie-loss tables we obtained when comparing the RB, MSE, ER, CR evaluation measures obtained by the different models over each outcome variable and by using $m = 42$ and $m = 5$ imputations are reported in, respectively, the Supplementary files S1, S2.

Observing the results, it is clear that in our case study IPW models systematically obtain less valid results. This is probably due to the fact that IPW can only use fully observed variables in the model for estimating the missingness probability, whereas MI uses all variables to estimate the conditional probabilities from which imputations are drawn. As a result, when the missingness pattern is complex and several variables contain missing values, as it is often the case in EHR data, there may be a lot of variables that thus cannot be used in the estimation of the inverse probability weights [16].

---

[18] In our model, variable binarization was performed by one-hot-encoding categorical, and binned numeric variables; this allows obtaining a normalized dataset ensuring that the following statistical estimates are expressed in the same scale.

Moreover, the complex missingness patterns that characterize EHR data often result in many individuals with at least one missing value. In these cases, inverse probability weighting may exhibit extreme power loss as too many rows need to be dropped.

For what regards the comparison of the imputation algorithms, all but four missRanger models (with no pmm and considering the outcome variables) obtained negative RB values and corresponding ER measure lower than one, meaning that all the models but missRanger with no pmm and considering the outcome variables underestimated the logodds computed on the complete dataset (p-value $< 0.05$). When comparing the results achieved by the algorithms exploiting iterative univariate imputation (Mice and missRanger models), we noted that the visiting order had a slight impact only in the case of missRanger, where the order given by the decreasing number of missing values produced lower RB distributions when compared to Amelia, Mice, and other missRanger algorithm settings (see the average RB values across the three outcomes and the sum of the wins, ties, losses over the three outcomes in, respectively, Figure 8 and supplementary Figure S1 and the per-outcome, colored win-tie-loss tables in Supplementary file S1). The slight behavioral differences among the two imputation orders suggested that the iterative procedure effectively reaches convergence.

On the other hand, the usage of the outcome variables in the imputation models did have an effect on the resulting evaluation measures. Amelia, Mice-norm, and Mice-logreg all achieved better results (lower absolute RB, lowest MSE, p-value $< 0.05$, Figures 8, supplementary Figures S1 and S2, and Supplementary file S1) when the outcome variables were included in the imputation model. Mice under the default settings appeared more robust with respect to the inclusion of the outcome variables. The behavior of missRanger with respect to the inclusion of the outcome variables strongly depended on the usage of the pmm estimator. Indeed, when pmm was used, the inclusion of the outcome achieved better results (p-value $< 0.05$); when pmm was not used, the inclusion of the outcome variables produced worse results (p-value $< 0.05$). Summarizing, all algorithms that used parametric approaches were improved by inclusion of the outcome variables, while algorithms solely based on RF classifier models were biased by the inclusion of the outcome.

Regarding the coverage rates (CR), all the models but one (Amelia with no outcome variables and one-hot-encoded binned numeric variables, first row in Figure 8) obtained values greater than the nominal rate (0.95), with a confidence interval lower than that obtained on the true estimates (the ratio SE measures are always lower than one).

Considering the results achieved when the imputation models work on one-hot-encoded (binned) numeric variables, only Amelia models seemed to be strongly impacted by an increase in the absolute values of the RB measure and of the MSE values and a decrease in the standard error. On the other hand, missRanger showed lower absolute values of RB and MSE measures when one-hot encoding of categorical and binned numeric variables was performed (see Figures 8, supplementary Figures S1 and S2, and Supplementary file S1).

Overall, the missRanger algorithms using no pmm and one-hot-encoding both categorical and binned numeric variables produced the most reliable results; they also achieved the lowest average standard errors, even when compared to the standard error obtained on the complete dataset (as shown by the ratio SE values, p-value < 0.05, Figure 8 and Supplementary file S1). With regards to the two other algorithms, among all the tested Mice models, Mice-norm with outcome variables achieves the lowest (absolute values of the) RB and MSE values (Figures 8, Supplementary Figures S1 and S2), outperforming also all the Amelia models; for Amelia, the inclusion of the outcome variables produced the best results.

When analyzing the results obtained by using only $m = 5$ imputations (Supplementary file S2), the conclusions drawn from the comparative evaluation measures over the 44 MI algorithms were similar. However, when comparing the evaluation measures obtained by each model on the m= 42 versus m=5 imputations, we noted that an higher number of imputations guarantees a lower variability of the evaluation measures with respect to different algorithm specifications, such as the usage of one-hot-encoded variables or the inclusion/deletion of the outcome variables in the prediction model. In other words, the higher the number of imputations, the higher the stability of the MI algorithm. This is particularly true for the missRanger algorithm.

Having chosen the best settings for each of the three algorithms we then ran them on the full dataset by computing $m = 42$ imputations. In this way, we obtained the odds estimates shown in Figure 5 (Hospitalization event), Figure 6 (Invasive ventilation event), and Figure 7 (Cox survival estimate).

To show the practicability of our evaluation method on different missingness mechanisms, while also providing a further assessment of the compared algorithms, we also run experiments by simulating MCAR missingness. For coherence with the MAR experiments, we therefore randomly amputated the BMI, Race, and Ethnicity predictors to produce, respectively, the 30%, the 15%, and the 15% of missing values (chosen by sampling from the uniform distribution). While MAR amputation maintains the proportion of cases with at least one missing value unaltered (42% in our case study), MCAR amputation produces random missingness patterns, resulting in different proportions of cases with at least one missing value across the amputations, i.e. different numbers, $m$, of multiple imputations per dataset. In particular, on the average of the amputations we drew, the proportion of cases with at least one missing value was (average ± standard error) 49% ± 2%, which resulted in an higher number of multiple imputations with respect to the MAR experiments ($m = 49 \pm 2$ imputations [40], Appendix A). Moreover, considering that the relationships between missing and observed data are not guaranteed in MCAR, it may be more difficult for the data imputation algorithm to estimate the imputation values. Indeed, all the algorithms we experimented needed more iterations to reach convergence; however, the computed evaluation metrics were comparable to those computed for MAR amputations, and also the behavior of the different models was confirmed (Figure 9 and Supplementary file S3_MCAR). This may suggest that, in this specific case study, the subset of complete cases remaining after amputation may suffice to describe the underlying data distribution; in other words, independent from the data being MAR or MCAR, the analysis of the observed data allows estimating the latent underlying relationships between predictors to recover the missing information.

*Generalizability of the evaluation method to different scenarios*

The evaluation method and evaluation metrics we are proposing are generalizable to different datasets, contexts, and scenarios.
Of course, depending on the problem at hand, any researcher may decide to favour one evaluation metric with respect to the others. As an example, besides the case study presented in subsection Case study: associations between diabetic-patients descriptors and COVID-19 hospitalization events, we applied our evaluation method in a context where we wanted to estimate the effect of a primary exposure while adjusting for other variables that may act as confounds [18]. In this context, the evaluation process we applied was the same: a complete dataset was obtained by listwise deletion and was adjusted by inverse probability weighting. A statistical estimator was then used to estimate the gold-standard treatment-effect on the adjusted cohort. Next, for each MI algorithm under evaluation, we amputated the complete dataset multiple times and we analyzed each amputated dataset as described in [69] to obtain MI-estimates of the treatment-effect on adjusted cohorts. Averaging all the MI treatment-effect estimates across all the amputations we obtained the expected value of the MI treatment-effect estimates. We next computed the numeric evaluation measures proposed above to compare the gold-standard treatment effect estimates to the expected values of the MI treatment-effect estimates. In this context, we considered as the most important evaluation metrics the raw bias (whose absolute value we desired to be as little as possible), the ratio between the standard errors of the estimates (which we required to be as little as possible), and the coverage rate (which we required to be as high as possible). This is because in our context we wanted to identify the MI algorithms that could provide reliable inference for the treatment-effect. In other words, treatment-effects estimates as protective (or dangerous) on the gold-standard (complete) dataset, should also result as protective (or dangerous) when estimated by the MI algorithm.

In the aforementioned case, the treatment predictor variable was the crucial one, while the others were adjusted; therefore we simply observed and compared the evaluation measures computed for that predictor over different MI algorithms. When instead all the predictor variables (or a subset of variables) in the dataset are of interest, we may exploit the advantage provided by the fact that our evaluation method proposes to normalize the dataset, which allows obtaining estimates characterized by the same scale. Therefore, we can average each evaluation measure across the different predictors (of interest), therefore obtaining unique $RB$, $MSE$, $ER$, $CR$ values characterizing the estimates of interest computed by applying the specific imputation method to the specific outcome variable.

On the other hand, we observe that each MI algorithm is well characterized by the $RB$, $MSE$, $ER$, $CR$ vectors that contain, respectively, the $rb_i$, $mse_i$, $er_i$ and $cr_i$ measures for each predictor variable in the dataset. Therefore, given a measure of interest (e.g., RB), its vector can be seen as a distribution characterizing the specific MI algorithm. The distribution characterizing an MI algorithm can therefore be compared to the RB vector characterizing a second MI algorithm by the Wilcoxon paired rank sign test.

**Discussion and Conclusions**

In this work we presented a novel method to numerically evaluate (imputation) algorithms for handling missing data in the context of statistical analysis. The method is general and can be used in different research fields and on datasets containing heterogeneous types.

To show the practicability of our method we analyzed a large cohort of patients with type 2 diabetes infected by COVID-19. The dataset, which contained a MAR missingness affecting an high proportion (42%) of cases, was used in a previous work to perform a complete case analysis to identify associations between crucial predictors and three COVID-19 outcomes (hospitalization, invasive mechanical ventilation, death).
By using the dataset we conducted thorough investigations to both answer computational and statistical questions about MI techniques, by also comparing them to commonly-applied forms of IPW models, and to validate our previous clinical results [15].
From a computational and statistical point of view, we were interested in understanding the differences between different specifications of (univariate and multivariate) multiple imputation techniques exploiting either flexible machine-learning based approaches [42] or parametric approaches [50][52] as the core inference model.

At first, the comparison between complete-case IPW and imputation models showed that, in our use-case, MI techniques perform systematically better, probably due to the decrease in power caused by the high proportion of cases (42%) that must be dropped from the weighted, complete-case analysis performed by the (non-augmented) IPW algorithm, applied here as an illustration of non-MI techniques that also find frequent use. This allows researchers to make considerations about the amount of missingness above which imputation should be applied with caution, even when assuming MAR data is available. In these regards, no theoretical limit has been defined yet in literature, though the work from Jakobsen et al. [29] concentrates on methods for handling missing data in randomized clinical trials and, based both on an extensive literature study, group discussions, and experience-in-the-field, suggests avoiding imputation when the proportions of missing data are large on important variables. In particular, when the proportions of missingness in predictors exceed, for example, the 40% authors suggest to just report the results of the (eventually weighted) complete case analysis and then clearly discuss the resulting interpretative limitations of the trial results. Indeed, as also noted by Clark and Altman [70] if the proportions of missing data are large on crucial variables, then the obtained estimates may only be considered as hypothesis generating results.

We agreed with suggestions from both Jakobsen et al. [29] and Clark and Altman [70]; however, we considered that other factors are relevant in this context. Indeed, the cardinality of the available complete cases, the number of variables in the dataset, and the complexity of the problem at hand should also be considered when evaluating whether the proportion of cases with missing values is too high. On one side, when the overall number of cases is limited and the number of variables is high, even lower proportions of missingness (e.g., 20%) may raise concern because the available complete cases may not be enough to fully represent the underlying data distribution.

When we applied our evaluation method to our use-case dataset (that contains a large cohort of complete cases and a limited number of predictors) and compared the validity of estimates computed by IPW or multiple imputation algorithms, we showed that Jakobsen's et al. missingness thresholds (proportions of missingness as high as 30%-40%) may be accepted when the available complete cases carry enough information about the underlying (MAR) data structure.

In any case, when the proportions of missingness exceed values (e.g., the 20%) we would suggest using our evaluation method to check the validity of data imputation models.

Obviously, a rare exception to this problem would be if it is relatively certain that the data are MCAR, because of the ignorability of this type of missingness. However, especially in biomedical/clinical data, the certainty of MCAR data is rare.

Our comparative evaluation highlighted that MI models exploiting machine-learning techniques, in this case RFs, tend to obtain the most reliable estimates and the lowest standard errors. Further, they produced more reliable estimates when the outcome variables were not considered during the imputation. This behavior is opposite to models assuming an underlying multivariate distribution (FCS and JM models), which yielded results comparable to machine learning-based models when considering the outcome variable during imputation. Additionally, the standard errors they obtained were always higher, better resembling the natural uncertainty in the data. The usage of one-hot-encoded categorical or binned numerical variables had no impact in the overall performance.

The aforementioned results were confirmed when the number of imputations was reduced to m = 5 (default value suggested by [4]), though in this case different algorithm specifications had a greater impact on the obtained evaluation measures. This result may suggest that Von Hippel's rule of thumb is a good way to guarantee robustness with respect to different algorithm specifications. However, considering our previous research [11], we would strongly suggest starting with Rubin's default value and then analyzing the behavior of the imputation model as the number of imputations increases to a value that is at least as high as the percent of missing cases. This would allow a researcher to assess the robustness of the obtained inferences. Therefore, since the fuzzy imputation resulting from one-hot-encoding may cause bias due to incoherent imputations, when possible we suggest avoiding one-hot-encoding.

Summarizing, exploiting the proposed MI-evaluation framework, we observe that, on our specific problem, missRanger (with no pmm application, no outcome variables in the imputation model, and without one-hot-encoding of categorical or binned numeric variables) is the most reliable imputation model broadly-speaking. We applied this approach to impute the available samples m = 42 times and then estimate pooled odds ratios (Figure 6, Figure 7 and Figure 8) which were compared by experts to those reported in Wong et al. [15]. For the sake of visual comparison we also computed pooled estimates by using Mice-Norm (with outcome variables in the imputation model) and Amelia (with outcome variables in the imputation model).

From a clinical standpoint, the results we obtained validated those presented in Wong et al. [15], where we investigated the relationships between HbA1c, BMI, demographics, medications, and comorbidities and the severity of COVID-19 infection outcomes. In all analyses, the primary findings of the study that risk of hospitalization increased with worsening levels of glycemic control, but that the risk of death plateaued at HbA1c >8 and ventilator or extracorporeal membrane oxygenation use plateaued at HbA1c>9 remained consistent. The findings from the present study strengthen the robustness of the primary findings in Wong et al. [15] and reduce the risk of severe bias due to the removal of cases with missing values [3][32][33].

There were some minor differences from the original report in the comparison of the effect of covariates on the risk of death, namely HIV was associated with a statistically significant decrease and cancer with a statistically significant increase in death in the MI data sets. When modeling the odds ratio (ORs) for ventilation or Extracorporeal membrane oxygenation (ECMO), there was also a statistically significant increased risk with dementia or severe liver disease, a significantly decreased risk with Sodium-glucose Cotransporter-2 (SGLT2) inhibitors, and no significant difference with sulfonylurea use in the imputed data sets. There was a small but statistically significant increase of hospitalization with pulmonary or peripheral vascular disease seen in the analysis using the imputed data set that was not seen with the complete data.

Of note, while in this paper we detailed the application of our evaluation method on a specific EHR-based dataset and for solving a clinical research question that did not require patient adjustment (via matching or weighing), in [18] we applied it to evaluate different MI algorithms to estimate the effect of a primary exposure while adjusting for other variables that may act as confounds (subsection Generalizability of the evaluation method to different scenarios). This shows that our evaluation approach can be applied to a broad range of (clinical) datasets to compare different strategies and methods to handle missing data while performing statistical analysis.

The evaluation measures described above provide a further advantage. In particular, we remind that in situations when the proportion of missingness is not so high to raise caution, a proper number of multiple imputations may be chosen based on the proportion of missingness (as detailed in Section A) to maximize the statistical efficiency of the MI estimator. As a result, the RB, MSE, ER, CR measures should only depend on the validity of the specific MI algorithm being used. On the other hand, when either the dataset is particularly complex, or it has high cardinality/dimensionality so that the computational costs of MI algorithms hamper the computation of a proper number of multiple imputations, the RB, MSE, ER, and CR measures could be worse for predictors whose imputation is problematic, due to their complexity and/or their high percentage of missingness. In this case, the analysis of the evaluation measures obtained by each predictor could be used to discard variables whose imputation is too problematic.

Our future work will be aimed at using the proposed evaluation method to perform a thorough comparison of different ways to deal with missing data when performing a statistical analysis. In particular, we will consider also cases when a single imputation strategy is preferred, by considering several context, scenarios, and datasets. As an example, we might consider interaction/effect modification cases, where, e.g., variables with missing data are multiplicative.

**Highlights**

- We propose an evaluation framework for comparing the validity of multiple imputation algorithms in a range of retrospective clinical studies where the need is to compute statistical estimates.

- While we focused on multiple imputation techniques, the generality of the method allows us to evaluate any missing-data handling strategy (e.g. IPW [16]), beyond those performing data imputation.

- The application of the evaluation method on a large cohort of patients from the N3C Enclave has shed some light of the following issues regarding the application of MI algorithms:

    - *inclusion of the outcome variable in the imputation model*: when choosing MI algorithms exploiting parametric univariate/multivariate estimators, the inclusion of the outcome variables in the imputations model can provide a better control for confounders. When applied to our clinical problem/dataset, MI algorithms exploiting estimators based on machine learning (RF) had the opposite behavior and tended to be biased by the inclusion of the outcome variables. However, caution is warranted here, as whether inclusion/exclusion of the outcome variable is best practice strongly depends on the properties of the data at hand. Indeed, the effect we observed on the output of RF-based models was likely due to the strong relationships between the outcome variables and the predictors. When using a different dataset where these relationships are stronger or weaker, the effect of the outcome inclusion may become either stronger or weaker, respectively. *Therefore, our "universal" guideline is to use our evaluation model to improve understanding of whether the outcome should be included for a particular application.*

    - *conversion of heterogeneous data types to homogeneous data types by one hot encoding*: when working on data types containing numeric and categorical predictors, some MI algorithms (e.g. Amelia and Mice with bayesian or pmm univariate imputation algorithm) abruptly convert categorical variables into numeric type, thus introducing a severe bias [51], and reducing the

validity of the obtained estimates. A solution to avoid this is to one-hot-encode categorical variables, therefore obtaining a set of binary predictors, whose scale and variability is however completely different from that of numeric variables. Testing under the setting where all variables (even numeric ones) are one-hot-encoded to obtain homogeneous predictors did not improve results. In particular, RF based MI algorithms seem the most stable with respect to data type heterogeneity, and this is due to their ability to handle mixed data types by design. *Therefore, when heterogeneous datasets must be treated, our "universal" guideline is to use imputation algorithms, e.g., RF-based methods, handling mixed data types by design.*

- *univariate imputation order*: when exploiting iterative univariate imputations (Mice and missRanger), the imputation order may have an impact on performance. The comparison of the imputation order according to either increasing or decreasing number of missing values showed only a slight impact on bias. *This implies that, under a reasonable number of iterations, the MI algorithms can reach convergence and the univariate imputation order may have no particular effect on the obtained results.* On the other hand, considering that we found no evidence in literature that allows to choose one order with respect to the other, when limited computational costs are available and hamper the computation of a large number of multiple imputations, *our "universal" guideline is to use our evaluation method to both check whether the univariate imputation order has any effect on the validity of the obtained estimates and, if this is the case, to also choose the most effective one.*

- *number of multiple imputations*: to choose the number of multiple imputations, Von Hippel's rule of thumb [40] would surely be a good choice. However, when dealing with large datasets, such a number of imputations can be prohibitive both from a time and memory perspective. For this reason, we would suggest performing a sensitivity analysis that starts with a low number of imputations (e.g. m=5 as suggested by Rubin [4]) and then proceeds towards large values until the evaluation measures stabilize. This process would also allow gaining additional insights about the behavior of different algorithms.

# ABBREVIATIONS

EHR: Electronic Health Record

MI: Multiple Imputation

LR: Logistic Regression (model)

CS: Cox-survival (model)

BMI: Body Mass Index

N3C: National COVID Cohort Collaborative

MCAR: Missing Completely At Random

MAR: Missing At Random

MNAR: Missing Not At Random

FCS: Fully Conditional Specification

JM: Joint Modeling

MIDA: Multiple Imputation using Denoising Autoencoders

GAN: Generative Adversarial Networks

GAIN: Generative Adversarial Imputation Networks

CART: Classification And Regression Trees

RF: Random Forest (classifier)

pmm: predictive mean matching (model)

RB: Raw Bias

ER: Estimate Ratio

MSE: Mean Squared Error

SE: Standard Error

CR: Coverage Rate

OR: odds ratio

ECMO: Extracorporeal membrane oxygenation

SGLT2: Sodium-glucose Cotransporter-2

IPW: Inverse Probability Weighting

# TABLES

*Table 1: the variables in Wong's et al. dataset [15], their type and their representation in the logistic regression and Cox-survival model. Numeric variables (age, BMI, Hba1c) were grouped and one-hot-encoded, while categorical variables (Gender, Ethnicity, and Race) were one-hot-encoded. To avoid collinearities, when one-hot-encoding a predictor variable, the binary predictor representing the largest group was left out for reference (marked with "used for reference" in the table). For each predictor group, the table also reports the percentage of missing cases, if any.*

| Predictor Group and predictor type | Predictor | percentage of missing values | all cases |
|---|---|---|---|
| **Number of cases (%)** | | | **56123 (100%)** |
| Gender  One-hot-encoded categorical variable | Male | | 49% |
| | female (used for reference) | | 51% |
| Age  Grouped and one-hot-encoded numeric variable | | | 61.88 ± 0.06 [18,89] |
| | age < 40 | | 7% |
| | 40 ≤ age < 50 | | 11% |
| | 50 ≤ age < 60 | | 22% |
| | 60 ≤ age < 70 (used for reference) | | 28% |
| | 70 ≤ age < 80 | | 22% |
| | age ≥ 80 | | 10% |
| BMI  Grouped and one-hot-encoded numeric variable | | 29% | 33.25 ± 0.04 [12.13,79.73] |
| | BMI < 20 | | 1% |
| | 20 ≤ BMI < 25 | | 8% |
| | 25 ≤ BMI < 30 | | 18% |
| | 30 ≤ BMI < 35 (used for reference) | | 18% |
| | 35 ≤ BMI < 40 | | 12% |
| | BMI ≥ 40 | | 13% |
| Race  One-hot-encoded categorical variable | White (used for reference) | 15% | 55% |
| | Other | | 1% |
| | Black | | 26% |
| | Asian | | 3% |
| Ethnicity | Hispanic | 12% | 16% |
| One-hot-encoded categoric variable | Not hispanic (used for reference) | | 73% |
| Hba1c | | | 7.58 ± 0.01 [4.1,19.3] |
| Grouped and one-hot-encoded numeric variable | Hba1c < 6 | | 17% |
| | 6 ≤ Hba1c < 7 (used for reference) | | 30% |
| | 7 ≤ Hba1c < 8 | | 21% |
| | 8 ≤ Hba1c < 9 | | 12% |
| | 9 ≤ Hba1c < 10 | | 07% |
| | Hba1c ≥ 10 | | 12% |
| Comorbidities | MI | | 13% |
| | CHF | | 23% |
| Binary variables; | PVD | | 21% |
| 1 = has comorbidity | Stroke | | 17% |
| 0 = does not have comorbidity | Dementia | | 5% |
| | Pulmonary | | 31% |
| | liver mild | | 16% |
| | liver severe | | 3% |
| | Renal | | 30% |
| | Cancer | | 14% |
| | Hiv | | 1% |
| Treatments | Metformin | | 26% |
| | dpp4 | | 5% |
| Binary variables; | sglt2 | | 5% |
| 1 = has comorbidity | Glp | | 7% |
| 0 = does not have comorbidity | Tzd | | 1% |
| | Insulin | | 25% |
| | Sulfonylurea | | 9% |

*Table 2: The notations used in the paper.*

| Parameter Name | Meaning |
|---|---|
| $N$ | number of cases (sample points) |
| $d$ | number of predictors |
| $X \in R^{N \times d}$ | a dataset containing $N$ cases, each described by $d$ predictors |
| $q_i, var_i, se_i, ci_i$ | the estimate (its variance, standard error, and confidence interval) computed on $X \in R^{N \times d}$ by a statical estimator for the $i^{th}$ predictor variable ($for\ each\ i \in \{1, \dots, d\}$). |
| $Q = [q_i]$, $VAR = [var_i]$, $SE = [se_i]$, $CI = [ci_i]$, for each $i \in \{1, \dots, d\}$ | The vector of all the estimates (their variance, standard error, and confidence interval) computed over all the predictors in a dataset $X \in R^{N \times d}$ |
| $m$ | number of multiple imputations |
| $\widehat{X^{(j)}} \in R^{N \times d}$ | The j-th imputed set |
| $\widehat{q_i^{(j)}}, \widehat{var_i^{(j)}}, \widehat{se_i^{(j)}}, \widehat{ci_i^{(j)}}$ for each $i \in \{1, \dots, d\}$ | the estimate (its variance, standard error, and confidence interval) for the $i^{th}$ predictor ($i \in \{1, \dots, d\}$) of the the j-th imputed set $\widehat{X^{(j)}} \in R^{N \times d}$ |
| $\widehat{Q^{(j)}} = \left[\widehat{q_i^{(j)}}\right]$ $\widehat{VAR^{(j)}} = \left[\widehat{var_i^{(j)}}\right]$ $\widehat{SE^{(j)}} = \left[\widehat{se_i^{(j)}}\right]$ $\widehat{CI^{(j)}} = \left[\widehat{ci_i^{(j)}}\right]$ for each $i \in \{1, \dots, d\}$ | The vector of all the estimates (their variance, standard error, and confidence interval) computed over all the predictors in the the j-th imputed set $\widehat{X^{(j)}} \in R^{N \times d}$. |
| $\widehat{q}_i, \widehat{var}_i, \widehat{se}_i, \widehat{ci}_i$ for each $i \in \{1, \dots, d\}$ | the pooled estimate (its variance, standard error, and confidence interval) obtained by an MI strategy for the $i^{th}$ predictor variable in $X \in R^{N \times d}$ by applying Rubin's rule (Rubin et al 1987). |
| $\hat{Q} = \frac{1}{m}\sum_{j=1}^{m} \widehat{Q^{(j)}} =$ <br><br> $= [\hat{q_i}]$, for each $i \in \{1,\ldots,d\}$ | The vector of the pooled estimates (one estimate per predictor variable) computed by an MI imputation strategy using $m$ imputations |
| $\widehat{W} = \frac{1}{m}\sum_{j=1}^{m} \widehat{VAR^{(j)}} \approx W_\infty$ <br><br> $\widehat{W} = [\widehat{W_i}]$, for each $i \in \{1,\ldots,d\}$ | $\widehat{W}$ is the vector of within imputation variances obtained with $m$ imputations (one within imputation variance per predictor variable). <br><br> $\widehat{W}$ is an estimate of $W_\infty$, the true within imputation variance when $m \to \infty$ |
| $\hat{B} = \frac{1}{m-1}\sum_{j=1}^{m} \left(\widehat{Q^{(j)}} - \hat{Q}\right)^2 \approx B_\infty$ <br><br> $\hat{B} = [\hat{B_i}]$, for each $i \in \{1,\ldots,d\}$ | $\hat{B}$ is the vector of between imputation variances obtained with $m$ imputations (one within imputation variance per predictor variable). <br><br> $\hat{B}$ is an estimate of $B_\infty$, the true between imputation variance when $m \to \infty$ |
| $\hat{T} = \widehat{W} + (1+\frac{1}{m})\hat{B} \approx T_\infty$ | $\hat{T}$ is the total variance that estimates the true total variance, $T_\infty$ when $m \to \infty$ |
| $A$ | The number of amputations of the complete dataset |
| $\underline{Q} = \frac{1}{A}\sum_{a=1}^{A} \widehat{Q(a)} = [\underline{q_i}] \approx E[\hat{Q}]$ <br> for each $i \in \{1,\ldots,d\}$ | The vector with the averages of the MI estimates across all the amputations, that approximates the (vector of) expected values of the MI estimates for each predictor |

*Table 3: MI algorithms, their (default and experimented) settings and the advantages and drawbacks evidenced by our experiments. Overall, we compared 44 different MI algorithms. They are four different specifications of Mice-default and Mice-logreg (using/avoiding the outcome variables in the imputation model and trying the imputation order given by the increasing/decreasing number of missing values), eight different specifications of MIce-Norm (where we also compared the usage of numeric variables - BMI/hba1c/age - versus the imputation and usage of one-hot-encoded binned numeric variables), four specifications of Amelia (using/avoiding the outcome variables in the MI model and using/one-hot encoding binned numeric variables) and 24 different specifications of missRanger (using/avoiding the outcome variables in the MI model and using/one-hot encoding binned numeric and categorical variables, using the imputation order provided by the increasing/decreasing order of missing values, and testing three different options for the pmm donors).*

| MI algorithm ($m = 5, 42$) | Mice-default (4 different specifications) | Mice-norm (8 different specifications) | Mice-logreg (4 different specifications) | missRanger | Amelia (4 different specifications) |
|---|---|---|---|---|---|
| **Univariate / multivariate imputation model** | univariate imputation by: pmm (continuous predictor) LR (binary predictors) polR (categorical predictors) | univariate imputation by Bayesian estimator | univariate imputation by LR | univariate imputation via RF | multivariate estimation of the distribution underlying the observed data via EM |
| **Univariate imputation order** | Increasing number of missing values (monotone order) Decreasing number of missing values (Reverse monotone order) | | | | Multivariate imputation model |
| **Use outcomes in the imputation model** | TRUE / FALSE | | | | |
| **One-hot-encoding of categorical predictors** | FALSE (default) | TRUE (default) | TRUE (default) | FALSE (default) TRUE | TRUE (default) |
| **One-hot-encoding of binned numeric predictors** | FALSE (default) | FALSE (default) TRUE | TRUE (default) | FALSE (default) TRUE | FALSE (default) TRUE |
| **pmm donors** | 3 donors | - | - | no pmm, 3 donors, 5 donors | |
| **Maximum number of iterations** | 21 | | | | |
| **notable ADVANTAGES and DRAWBACKS** | ADVANTAGES:<br><br>1) usage of ad-hoc univariate imputation models based on predictor type<br><br>2) collinearities in predictor data are detected and reported to allow users to repair the problem | ADVANTAGES:<br><br>1) collinearities in predictor data are detected and reported to allow users to repair the problem | ADVANTAGES:<br><br>1) collinearities in predictor data are detected and reported to allow users to repair the problem | ADVANTAGES:<br>1) deals with heterogeneous data types<br><br>2) low variance when predictive mean matching is not used<br><br>3) application of pmm avoids the generations of values outside the original data distribution<br><br>DRAWBACKS:<br>1) RFs may take lot of iterations to converge when not informative predictors are provided | ADVANTAGES: 1) identifies collinearities that may alter results<br><br>2) faster than Mice and missRanger when working on datasets having large cardinalities<br><br>DRAWBACKS:<br>1) when data collinearities are detected, the predictors causing the collinearities are not reported. In this case, the matrix is singular and Amelia crashes |

*Table 4: dataset statistics for all the 56,123 patients (column "all cases"), hospitalized patients (column "hospitalized cases") and non-hospitalized patients (column "non-hospitalized cases"). Column p-value reports, for each predictor, the p-value for the null hypothesis of a Pearson correlation test between the binary predictor and the outcome variable (Hospitalization event).*

| Hospitalization event | | | | | | |
|---|---|---|---|---|---|---|
| **Predictor Group** | **Predictor** | **% of missing values** | **all cases** | **hospitalized cases** | **non-hospitalized cases** | **p-value** |
| **Number of cases (%)** | | | 56123 (100%) | 25399 (45%) | 30725 (55%) | |
| Gender | Male | | 49% | 23% | 26% | < 0.0001 |
| | Female | | 51% | 22% | 29% | < 0.0001 |
| Age | | | 61.88 ± 0.06 [18,89] | 63.7 ± 0.09 [18,89] | 60.39 ± 0.08 [18,89] | < 0.0001 |
| | age < 40 | | 7% | 3% | 4% | < 0.0001 |
| | 40 ≤ age < 50 | | 11% | 4% | 7% | < 0.0001 |
| | 50 ≤ age < 60 | | 22% | 9% | 14% | < 0.0001 |
| | 60 ≤ age < 70 (used for reference) | | 28% | 13% | 15% | ~ 0.4257 |
| | 70 ≤ age < 80 | | 22% | 11% | 11% | < 0.0001 |
| | age ≥ 80 | | 10% | 6% | 4% | < 0.0001 |
| BMI | | | 33.25 ± 0.04 [12.13,79.73] | 32.87 ± 0.06 [12.4,79.73] | 33.62 ± 0.06 [12.13,78.98] | < 0.0001 |
| | BMI < 20 | | 1% | 1% | 0% | < 0.0001 |
| | 20 ≤ BMI < 25 | | 8% | 5% | 4% | < 0.0001 |
| | 25 ≤ BMI < 30 | 29% | 18% | 9% | 9% | ~ 0.0104 |
| | 30 ≤ BMI < 35 (used for reference) | | 18% | 8% | 10% | < 0.0001 |
| | 35 ≤ BMI < 40 | | 12% | 5% | 7% | < 0.0001 |
| | BMI ≥ 40 | | 13% | 6% | 7% | ~ 0.0574 |
| Race | White (used for reference) | | 55% | 23% | 32% | < 0.0001 |
| | Other | 15% | 1% | 0% | 1% | ~ 0.1463 |
| | Black | | 26% | 14% | 12% | < 0.0001 |
| | Asian | | 3% | 1% | 2% | ~ 0.0032 |
| Ethnicity | Hispanic | 12% | 16% | 8% | 8% | < 0.0001 |
| | Not hispanic (used for reference) | | 73% | 33% | 40% | < 0.0001 |
| Hba1c | | | 7.58 ± 0.01 [4.1,19.3] | 7.78 ± 0.01 [4.1,19.3] | 7.41 ± 0.01 [4.1,18.7] | < 0.0001 |
| | Hba1c < 6 | | 17% | 8% | 9% | ~ 0.4878 |
| | 6 ≤ Hba1c < 7 (used for reference) | | 30% | 12% | 18% | < 0.0001 |
| | 7 ≤ Hba1c < 8 | | 21% | 09% | 12% | < 0.0001 |
| | 8 ≤ Hba1c < 9 | | 12% | 06% | 6% | < 0.0001 |
| | 9 ≤ Hba1c < 10 | | 07% | 04% | 3% | < 0.0001 |
| | Hba1c ≥ 10 | | 12% | 07% | 5% | < 0.0001 |
| Comorbidities | MI | | 13% | 8% | 5% | < 0.0001 |
| | CHF | | 23% | 14% | 9% | < 0.0001 |
| | PVD | | 21% | 12% | 9% | < 0.0001 |
| | Stroke | | 17% | 10% | 7% | < 0.0001 |
| | Dementia | | 5% | 3% | 2% | < 0.0001 |
| | Pulmonary | | 31% | 15% | 16% | < 0.0001 |
| | liver_mild | | 16% | 7% | 8% | < 0.0001 |
| | liver_severe | | 3% | 2% | 01% | < 0.0001 |
| | Renal | | 30% | 18% | 12% | < 0.0001 |
| | Cancer | | 14% | 7% | 07% | < 0.0001 |
| | Hiv | | 1% | 0% | 1% | ~ 0.5856 |
| Treatments | Metformin | | 26% | 8% | 18% | < 0.0001 |
| | dpp4 | | 5% | 2% | 3% | < 0.0001 |
| | sglt2 | | 5% | 2% | 3% | < 0.0001 |
| | Glp | | 7% | 2% | 5% | < 0.0001 |
| | Tzd | | 1% | 0% | 1% | < 0.0001 |
| | Insulin | | 25% | 14% | 11% | < 0.0001 |
| | Sulfonylurea | | 9% | 03% | 6% | < 0.0001 |

*Table 5: dataset statistics for all the 56,123 patients (column "all cases"), patients who were treated with invasive ventilation (column "cases with invasive ventilation") and were not treated with invasive ventilation (column "cases without invasive ventilation"). Column p-value reports, for each predictor, the p-value for the null hypothesis of a Pearson correlation test between the binary predictor and the outcome variable (Invasive ventilation event).*

| Invasive ventilation event | | | | | | |
|---|---|---|---|---|---|---|
| **Predictor Group** | **predictor** | **% of missing values** | **all cases** | **cases with invasive ventilation** | **cases without invasive ventilation** | **p-value** |
| **Number of cases (%)** | | | 56123 (100%) | 3623 (6.6%) | 52500 (93.4%) | |
| Gender | male | | 49% | 4% | 0.45% | < 0.0001 |
| | female | | 51% | 3% | 0.48% | < 0.0001 |
| Age | | | 61.88 ± 0.06 [18,89] | 64.22 ± 0.21 [19,89] | 61.72 ± 0.06 [18,89] | < 0.0001 |
| | age < 40 | | 7% | 0% | 7% | < 0.0001 |
| | 40 ≤ age < 50 | | 11% | 0% | 11% | < 0.0001 |
| | 50 ≤ age < 60 | | 22% | 1% | 21% | ~ 2e-04 |
| | 60 ≤ age < 70 (used for reference) | | 28% | 2% | 26% | < 0.0001 |
| | 70 ≤ age < 80 | | 22% | 2% | 20% | < 0.0001 |
| | age ≥ 80 | | 10% | 1% | 9% | ~ 0.4262 |
| BMI | | 29% | 33.25 ± 0.04 [12.13,79.73] | 33.44 ± 0.18 [13,77.41] | 33.24 ± 0.04 [12.13,79.73] | ~ 0.5314 |
| | BMI < 20 | | 1% | 0% | 0% | ~ 2e-04 |
| | 20 ≤ BMI < 25 | | 8% | 1% | 7% | ~ 0.496 |
| | 25 ≤ BMI < 30 | | 18% | 1% | 17% | ~ 0.538 |
| | 30 ≤ BMI < 35 (used for reference) | | 18% | 1% | 17% | ~ 0.6923 |
| | 35 ≤ BMI < 40 | | 12% | 1% | 11% | ~ 0.1574 |
| | BMI ≥ 40 | | 13% | 1% | 12% | ~ 0.5489 |
| Race | White (used for reference) | 15% | 55% | 3% | 52% | < 0.0001 |
| | Other | | 1% | 0% | 1% | ~ 1e-04 |
| | Black | | 26% | 2% | 24% | < 0.0001 |
| | Asian | | 3% | 0% | 3% | ~ 0.1976 |
| Ethnicity | Hispanic | | 16% | 1% | 15% | < 0.0001 |
| | Not hispanic (used for reference) | 12% | 73% | 5% | 68% | < 0.0001 |
| Hba1c | | | 7.58 ± 0.01 [4.1,19.3] | 7.83 ± 0.04 [4.1,18.3] | 7.56 ± 0.01 [4.1,19.3] | < 0.0001 |
| | Hba1c < 6 | | 17% | 1% | 16% | ~ 0.3225 |
| | 6 ≤ Hba1c < 7 (used for reference) | | 30% | 2% | 29% | < 0.0001 |
| | 7 ≤ Hba1c < 8 | | 21% | 1% | 20% | ~ 0.1481 |
| | 8 ≤ Hba1c < 9 | | 12% | 1% | 11% | < 0.0001 |
| | 9 ≤ Hba1c < 10 | | 7% | 1% | 6% | < 0.0001 |
| | Hba1c ≥ 10 | | 12% | 1% | 11% | < 0.0001 |
| Comorbidities | MI | | 13% | 1% | 12% | < 0.0001 |
| | CHF | | 23% | 2% | 21% | < 0.0001 |
| | PVD | | 21% | 2% | 19% | < 0.0001 |
| | stroke | | 17% | 1% | 16% | < 0.0001 |
| | dementia | | 5% | 0% | 5% | ~ 0.269 |
| | pulmonary | | 31% | 2% | 29% | ~ 0.0013 |
| | liver_mild | | 16% | 1% | 15% | ~ 0.4166 |
| | liver_severe | | 3% | 0% | 3% | < 0.0001 |
| | renal | | 30% | 3% | 27% | < 0.0001 |
| | cancer | | 14% | 1% | 13% | ~ 0.0186 |
| | hiv | | 1% | 0% | 1% | ~ 0.6854 |
| Treatments | metformin | | 26% | 1% | 25% | < 0.0001 |
| | dpp4 | | 5% | 0% | 5% | < 0.0001 |
| | sglt2 | | 5% | 0% | 5% | < 0.0001 |
| | glp | | 7% | 0% | 7% | < 0.0001 |
| | tzd | | 1% | 0% | 1% | ~ 0.0272 |
| | insulin | | 25% | 2% | 23% | < 0.0001 |
| | sulfonylurea | | 9% | 0% | 9% | < 0.0001 |

*Table 6: dataset statistics for all the 56,123 patients (column "all cases"), patients who died (column "death") and who survived (column "survival"). Column p-value reports, for each predictor, the p-value for the null hypothesis of a Pearson correlation test between the binary predictor and the outcome variable (death event).*

| Death event | | | | | | |
|---|---|---|---|---|---|---|
| **Predictor Group** | **predictor** | **% of missing values** | **sample distribution** | **death** | **Survival** | **p-value** |
| **Number of cases (%)** | | | **56123 (100%)** | **2865 (5.1%)** | **53258 (94.9%)** | |
| Gender | male | | 49% | 3% | 46% | < 0.0001 |
| | female | | 51% | 2% | 49% | < 0.0001 |
| Age | | | 61.88 ± 0.06 [18,89] | 71.34 ± 0.21 [20,89] | 61.38 ± 0.06 [18,89] | < 0.0001 |
| | age < 40 | | 7% | 0% | 7% | < 0.0001 |
| | 40 ≤ age < 50 | | 11% | 0% | 11% | < 0.0001 |
| | 50 ≤ age < 60 | | 22% | 1% | 22% | < 0.0001 |
| | 60 ≤ age < 70 (used for reference) | | 28% | 1% | 27% | < 0.0001 |
| | 70 ≤ age < 80 | | 22% | 2% | 20% | < 0.0001 |
| | age ≥ 80 | | 10% | 1% | 9% | < 0.0001 |
| BMI | | | 33.25 ± 0.04 [12.13,79.73] | 31.47 ± 0.18 [12.4,74.91] | 33.36 ± 0.04 [12.13,79.73] | < 0.0001 |
| | BMI < 20 | 29% | 1% | 0% | 0% | < 0.0001 |
| | 20 ≤ BMI < 25 | | 8% | 1% | 7% | < 0.0001 |
| | 25 ≤ BMI < 30 | | 18% | 1% | 17% | ~ 0.0029 |
| | 30 ≤ BMI < 35 (used for reference) | | 18% | 1% | 17% | ~ 0.0114 |
| | 35 ≤ BMI < 40 | | 12% | 0% | 12% | ~ 7e-04 |
| | BMI ≥ 40 | | 13% | 1% | 12% | < 0.0001 |
| Race | White (used for reference) | | 55% | 3% | 52% | ~ 0.0452 |
| | Other | 15% | 1% | 0% | 1% | ~ 0.77 |
| | Black | | 26% | 1% | 25% | ~ 0.0385 |
| | Asian | | 3% | 0% | 3% | ~ 0.8894 |
| Ethnicity | Hispanic | 12% | 16% | 1% | 15% | ~ 0.6093 |

| | | | | | |
|---|---|---|---|---|---|
| | Not hispanic (used for reference) | | 73% | 0.04% | 0.69% | ~ 0.6093 |
| Hba1c | | | 7.58 ± 0.01 [4.1,19.3] | 7.55 ± 0.04 [4.2,18] | 7.58 ± 0.01 [4.1,19.3] | ~ 0.3272 |
| | Hba1c < 6 | | 17% | 1% | 16% | ~ 0.833 |
| | 6 ≤ Hba1c < 7 (used for reference) | | 30% | 1% | 29% | ~ 0.0522 |
| | 7 ≤ Hba1c < 8 | | 21% | 1% | 20% | ~ 0.2162 |
| | 8 ≤ Hba1c < 9 | | 12% | 1% | 12% | < 0.0001 |
| | 9 ≤ Hba1c < 10 | | 07% | 0% | 7% | ~ 0.3081 |
| | Hba1c ≥ 10 | | 12% | 1% | 11% | ~ 6e-04 |
| Comorbidities | MI | | 13% | 1% | 12% | < 0.0001 |
| | CHF | | 23% | 2% | 21% | < 0.0001 |
| | PVD | | 21% | 2% | 19% | < 0.0001 |
| | stroke | | 17% | 1% | 16% | < 0.0001 |
| | dementia | | 5% | 1% | 4% | < 0.0001 |
| | pulmonary | | 31% | 2% | 29% | < 0.0001 |
| | liver_mild | | 16% | 1% | 15% | ~ 0.7307 |
| | liver_severe | | 3% | 0% | 3% | < 0.0001 |
| | renal | | 30% | 3% | 27% | < 0.0001 |
| | cancer | | 14% | 1% | 13% | < 0.0001 |
| | hiv | | 1% | 0% | 1% | ~ 0.0048 |
| Treatments | metformin | | 26% | 1% | 25% | < 0.0001 |
| | dpp4 | | 5% | 0% | 5% | ~ 0.0217 |
| | sglt2 | | 5% | 0% | 5% | < 0.0001 |
| | glp | | 7% | 0% | 7% | < 0.0001 |
| | tzd | | 1% | 0% | 1% | ~ 0.0029 |
| | insulin | | 25% | 2% | 23% | < 0.0001 |
| | sulfonylurea | | 9% | 0% | 9% | < 0.0001 |

# FIGURES

| Ethnicity | Race | BMI | no cases with missingness pattern | Percentage w.r.t. the number of cases | # variables composing the missing pattern |
|---|---|---|---|---|---|
| | | | 56123 | 100.0% | Total number of samples |
| | | | 32529 | 58% | Complete observations |
| | | | 23594 | 42% | Cases with missing values |
| | | BMI missing | 9993 | 17.8% | cases where only BMI is missing |
| | Race missing | | 4951 | 8.8% | cases where only Race is missing |
| Ethnicity missing | | | 1226 | 2.2% | cases where only Ethnicity is missing |
| | Race missing | BMI missing | 2159 | 3.8% | cases where Race and BMI are missing |
| Ethnicity missing | | BMI missing | 3732 | 6.6% | cases where Ethnicity and BMI are missing |
| Ethnicity missing | Race missing | | 910 | 1.6% | cases where Ethnicity and Race are missing |
| Ethnicity missing | Race missing | BMI missing | 623 | 1.1% | cases where Ethnicity, Race, and BMI are missing |
| 6491 | 8643 | 16507 | | | |
| 11.6% | 15.4% | 29.4% | | | |

Figure 1: The presumed MAR missing data patterns in the Wong's et al. [15] dataset.

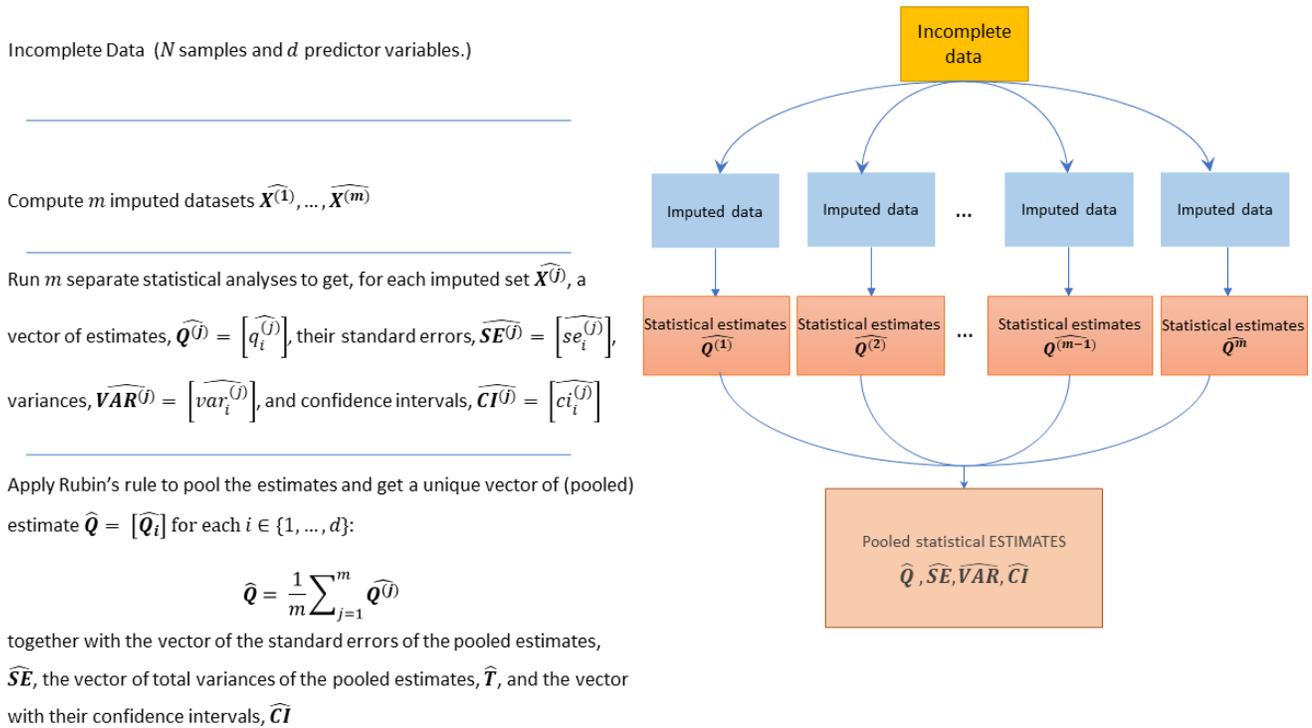

Figure 2: schematic diagram of the pipeline used to obtain pooled estimates when applying a MI strategy. The incomplete dataset is imputed m times, where the value of m can be defined in order to maximize the efficiency of the MI estimator (see Appendix A); each imputed dataset is individually processed to compute separate inferences; all the inferences are pooled by Rubin's rule [4] to get the pooled estimates ($\widehat{Q}$), their total variances ($\widehat{VAR}$) and standard errors ($\widehat{SE}$) and their confidence intervals ($\widehat{CI}$). In the figure, we use the superscript j to index the imputations number (j $\in \{1, ..., m\}$) and the subscript i to index the predictor variable in the dataset (see Table 2 for a detailed list of all the notations used throughout the paper).

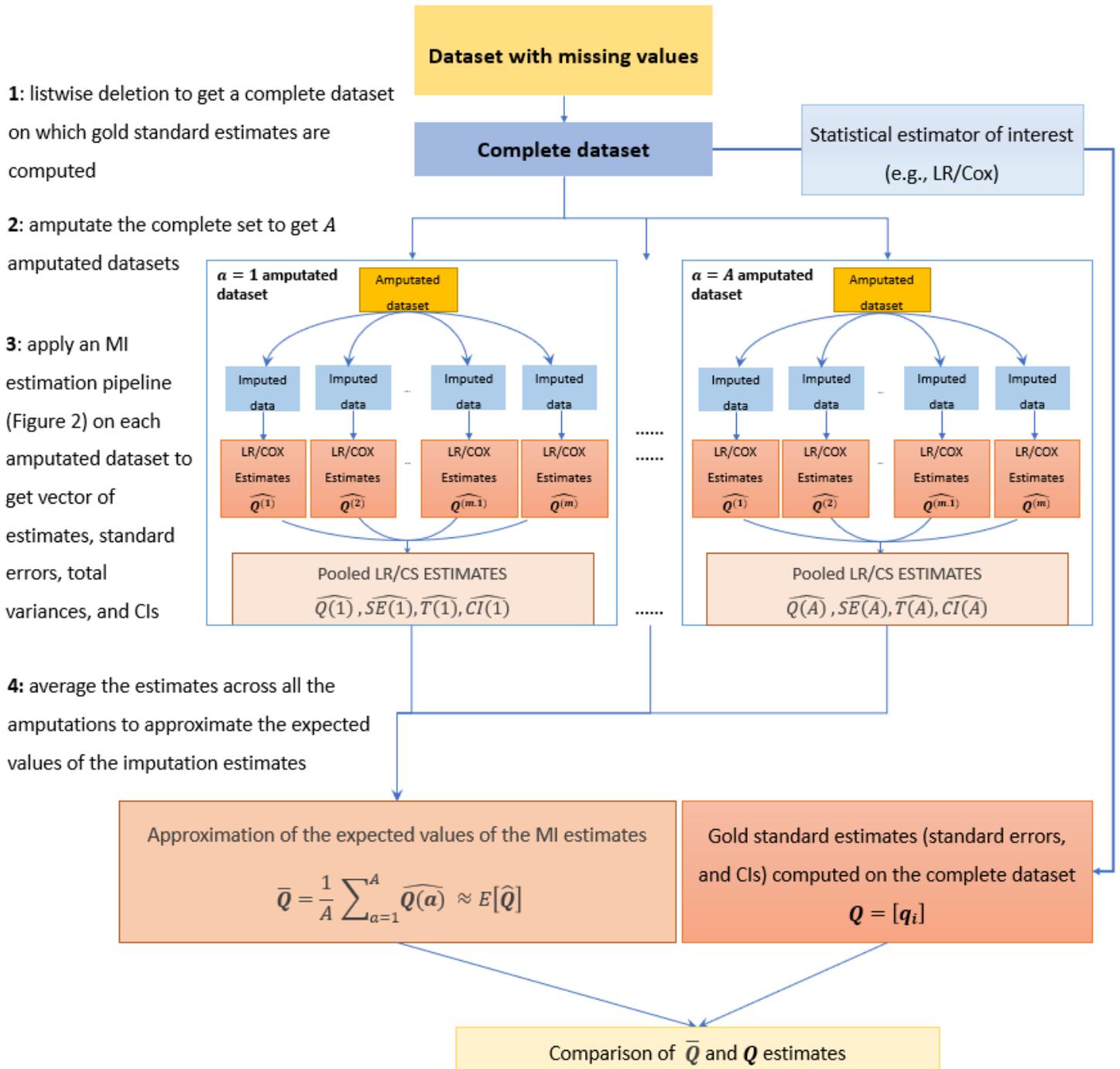

Figure 3: schematic diagram of the pipeline used to evaluate one MI algorithm across A multiple amputation settings. The following steps are applied: 1) listwise deletion is used to produce a complete dataset on which a vector of estimates to be used as "gold standard" is computed; 2) a number A of amputated datasets is computed by using an amputation algorithm that reproduces the same missingness pattern in the original dataset; 3) An MI estimation pipelines (see Figure 2) are applied to get A pooled estimates, their total variances, standard errors and confidence intervals; 4) averaging the A estimates the expected value of the MI estimates are approximated and compared to the gold standard estimates computed on the complete dataset (step 1).

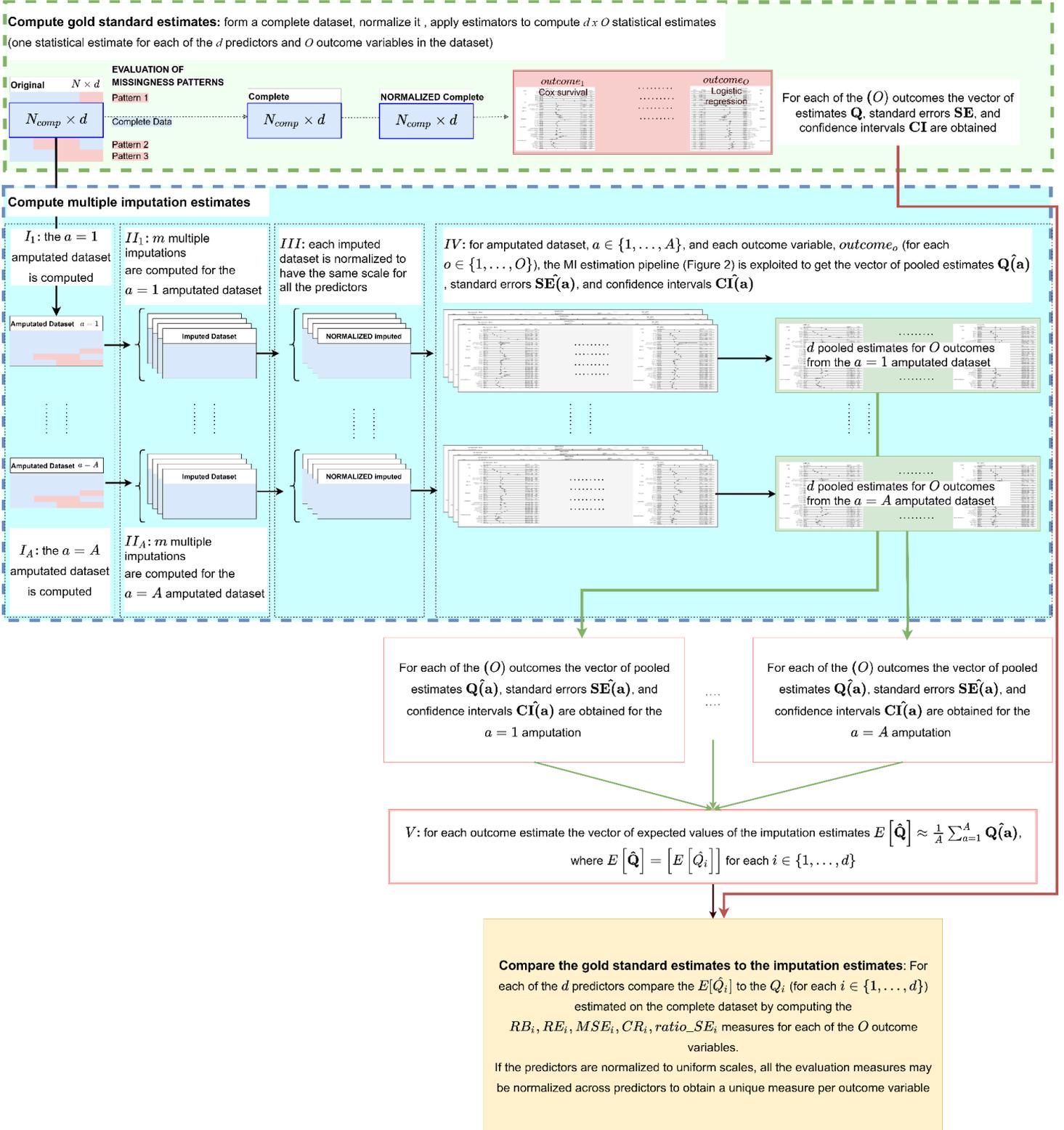

Figure 4: schematic diagram of the pipeline used to evaluate a multiple imputation algorithm. [**TOP light green BOX**: compute gold standard estimates] Listwise deletion is used to create a "complete" dataset where all the values are observed; the predictor variables in the complete dataset are normalized to obtain uniform scales across different predictors; statistical estimators (in our experiments they were two logistic regression models and one Cox survival model) are applied to compute statistical estimates describing the influence of the available predictors on O outcome variables (in our experiments they were O = 3 outcomes describing the hospitalization event, the invasive ventilation event, and patients' survival). [**BOTTOM light blue BOX**: compute MI estimates] ($I_1, ..., I_A$) From the complete dataset, A amputated dataset are computed; ($II_1, ..., II_A$) each amputated dataset is imputed m times by the MI algorithm under evaluation and (III) each imputed dataset is normalized (as done in the TOP BOX for the complete dataset) to obtain uniform scales across all the predictors in all the imputed datasets and in the complete dataset. (IV) Each imputed-normalized dataset is processed by the O statistical estimators and Rubin's rule [4] is applied to pool the estimates across the m imputations. (V) The pooled estimates obtained for each outcome and predictor variable are averaged across the A simulations (1 simulation per amputated dataset) to approximate the expected values of the estimate for each predictor and outcome. [**YELLOW BOX: compare the gold standard estimates to the imputation estimates**] The evaluation measures detailed in Section "Evaluation method" are computed for comparing the computed estimates to the gold standard estimates computed on the complete-normalized dataset (for each of the predictors and outcome variables). Of note, the normalization of the (complete and imputed) dataset predictors to a unique scale before the estimation would allow averaging all the evaluation measures across all the predictors.

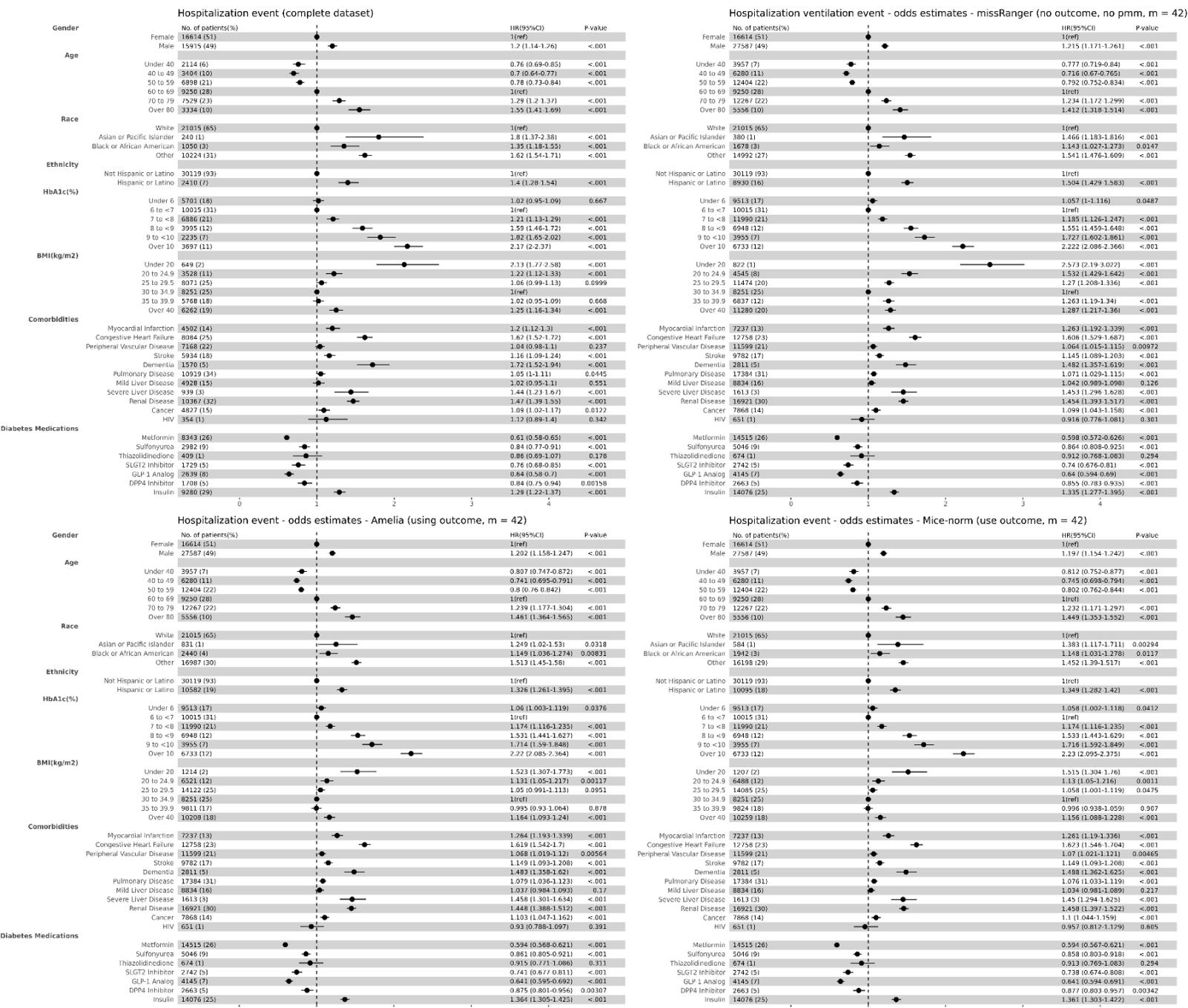

Figure 5: Hospitalization event: estimates obtained on the complete dataset obtained by listwise deletion (top-left) and on the full dataset by the MI estimation pipelines that use the best missRanger (top-right), Amelia (bottom-left), and Mice (bottom-right) models. For missRanger we used no pmm, we did not use the outcome variables in the imputation model, we one-hot encoded categorical predictors and binned numeric predictors (age, BMI, and Hba1c), and we used an univariate imputation order given by the decreasing number of missing values; for Mice-norm we included the outcome variables in the imputation model, we used an univariate imputation order given by the increasing number of missing values; for Amelia we included the outcome variables in the imputation model.

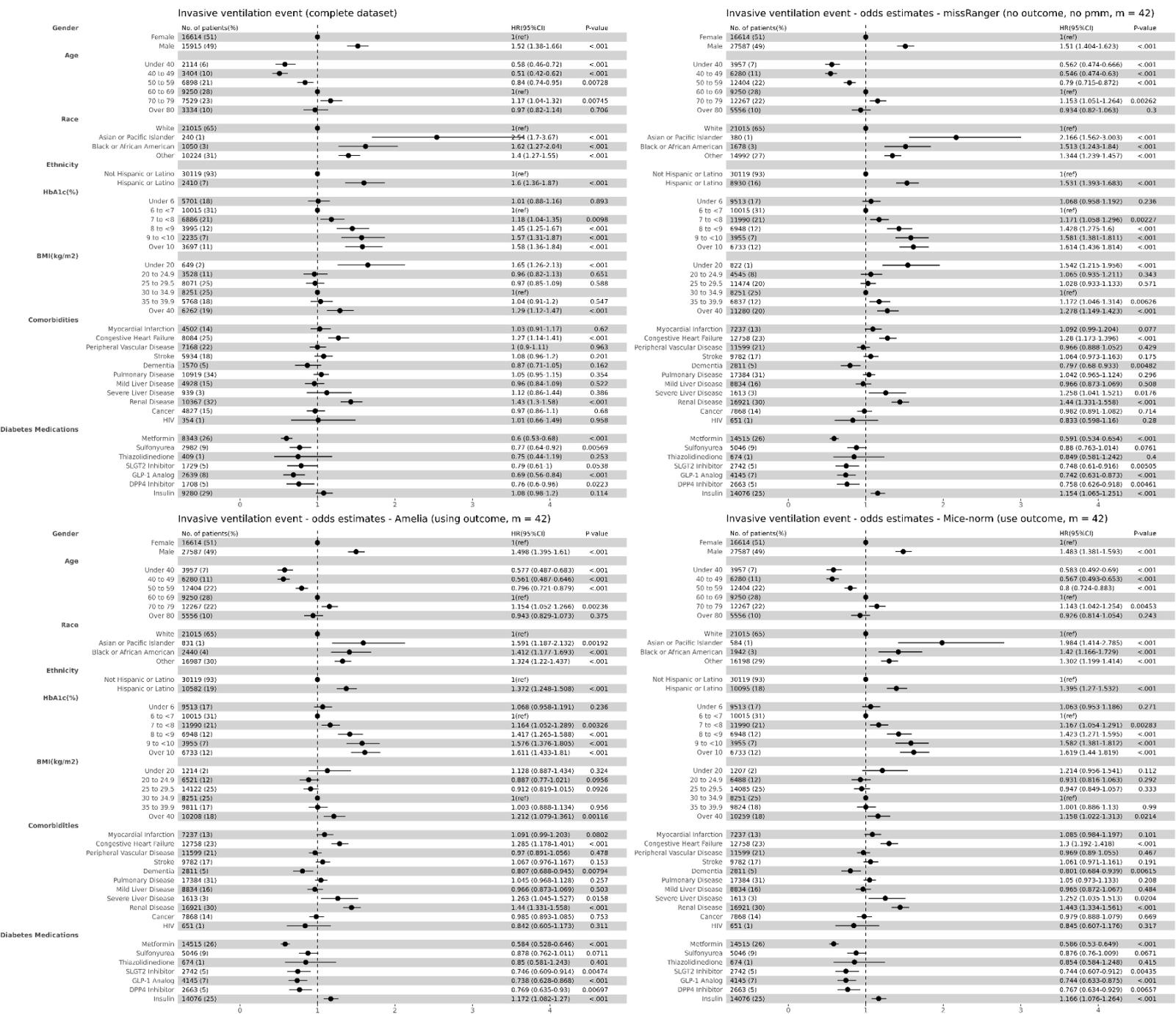

Figure 6: Invasive ventilation event: estimates obtained on the complete dataset obtained by listwise deletion (top-left) and on the full dataset by the MI estimation pipelines that using the best missRanger (top-right), Amelia (bottom-left), and Mice (bottom-right) models. For missRanger we used no pmm, we did not use the outcome variables in the imputation model, we one-hot encoded categorical predictors and binned numeric predictors (age, BMI, and Hba1c), and we used an univariate imputation order given by the decreasing number of missing values; for Mice-norm we included the outcome variables in the imputation model, we used an univariate imputation order given by the increasing number of missing values; for Amelia we included the outcome variables in the imputation model.

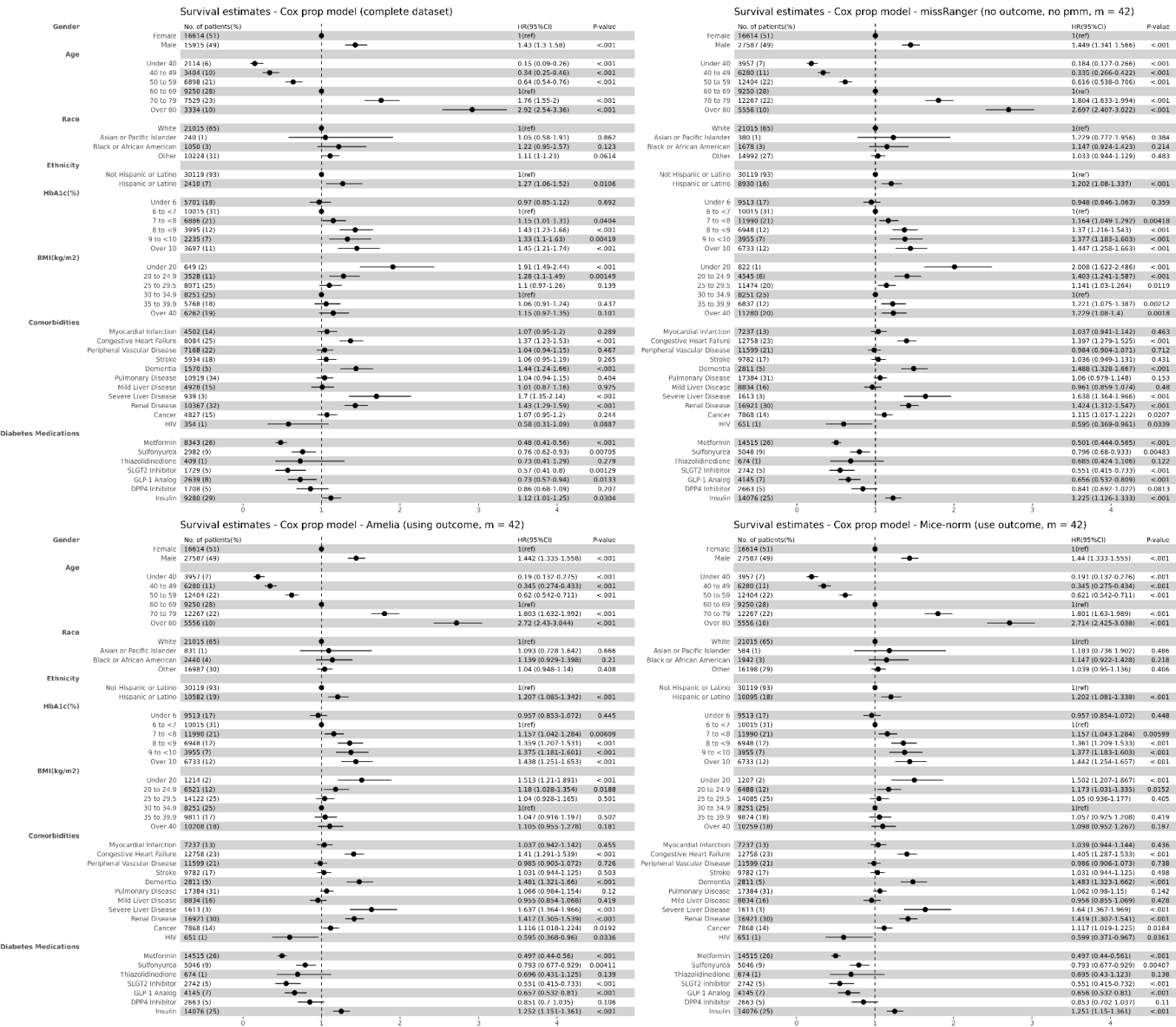

Figure 7: Death event: estimates obtained on the complete dataset obtained by listwise deletion (top-left) and on the full dataset by the MI estimation pipelines that using the best missRanger (top-right), Amelia (bottom-left), and Mice (bottom-right) models. For missRanger we used no pmm, we did not use the outcome variables in the imputation model, we one-hot encoded categorical predictors and binned numeric predictors (age, BMI, and Hba1c), and we used an univariate imputation order given by the decreasing number of missing values; for Mice-norm we included the outcome variables in the imputation model, we used an univariate imputation order given by the increasing number of missing values; for Amelia we included the outcome variables in the imputation model.

| MI algorithm | univariate imputation method | use outcomes | one-hot encode binned numeric predictors | one-hot encode categorical predictors | univariate imputation order | pmm donors | average of absolute values of RB across outcomes | average MSE across outcomes | average ER across outcomes | average ER across outcomes | average ER across outcomes |
|---|---|---|---|---|---|---|---|---|---|---|---|
| amelia | | F | T | | | | 0.025 | 0.007 | 0.900 | 0.939 | 0.722 |
| | | | F | T | | | 0.017 | 0.004 | 0.949 | 0.974 | 0.734 |
| | | T | T | | | | 0.021 | 0.005 | 0.914 | 0.966 | 0.724 |
| | | | F | | | | 0.014 | 0.003 | 0.971 | 0.988 | 0.735 |
| mice | default | F | F | F | monotone | | 0.012 | 0.002 | 0.962 | 0.981 | 0.737 |
| | | | | | revmonotone | | 0.012 | 0.002 | 0.964 | 0.981 | 0.737 |
| | | T | | | monotone | | 0.012 | 0.002 | 0.963 | 0.981 | 0.737 |
| | | | | | revmonotone | | 0.012 | 0.002 | 0.962 | 0.980 | 0.737 |
| | logreg | F | T | T | monotone | | 0.013 | 0.003 | 0.950 | 0.979 | 0.736 |
| | | | | | revmonotone | | 0.013 | 0.003 | 0.950 | 0.978 | 0.736 |
| | | T | | | monotone | | 0.011 | 0.002 | 0.952 | 0.987 | 0.736 |
| | | | | | revmonotone | | 0.011 | 0.002 | 0.952 | 0.986 | 0.736 |
| | norm | F | T | T | monotone | | 0.012 | 0.002 | 0.952 | 0.978 | 0.737 |
| | | | | | revmonotone | | 0.012 | 0.002 | 0.950 | 0.977 | 0.737 |
| | | | F | | monotone | | 0.012 | 0.002 | 0.963 | 0.980 | 0.737 |
| | | | | | revmonotone | | 0.013 | 0.002 | 0.961 | 0.980 | 0.737 |
| | | T | T | | monotone | | 0.011 | 0.002 | 0.952 | 0.986 | 0.737 |
| | | | | | revmonotone | | 0.011 | 0.002 | 0.951 | 0.987 | 0.737 |
| | | | F | | monotone | | 0.011 | 0.002 | 0.964 | 0.988 | 0.738 |
| | | | | | revmonotone | | 0.011 | 0.002 | 0.963 | 0.989 | 0.738 |
| missRanger | extratrees | F | T | T | monotone | | 0.007 | 0.001 | 0.952 | 0.994 | 0.733 |
| | | | | | monotone | 3 | 0.008 | 0.001 | 0.958 | 0.993 | 0.760 |
| | | | | | monotone | 5 | 0.009 | 0.001 | 0.956 | 0.992 | 0.757 |
| | | | | | revmonotone | | 0.007 | 0.001 | 0.951 | 0.993 | 0.733 |
| | | | | | revmonotone | 3 | 0.008 | 0.001 | 0.958 | 0.994 | 0.760 |
| | | | | | revmonotone | 5 | 0.009 | 0.001 | 0.955 | 0.992 | 0.757 |
| | | | F | F | monotone | | 0.007 | 0.001 | 0.955 | 0.996 | 0.730 |
| | | | | | monotone | 3 | 0.010 | 0.002 | 0.960 | 0.987 | 0.751 |
| | | | | | monotone | 5 | 0.010 | 0.002 | 0.961 | 0.983 | 0.746 |
| | | | | | revmonotone | | 0.007 | 0.001 | 0.956 | 0.996 | 0.730 |
| | | | | | revmonotone | 3 | 0.010 | 0.002 | 0.959 | 0.987 | 0.750 |
| | | | | | revmonotone | 5 | 0.010 | 0.002 | 0.960 | 0.983 | 0.747 |
| | | T | T | T | monotone | | 0.006 | 0.001 | 1.037 | 0.993 | 0.741 |
| | | | | | monotone | 3 | 0.006 | 0.001 | 0.971 | 1.000 | 0.785 |
| | | | | | monotone | 5 | 0.007 | 0.001 | 0.967 | 1.000 | 0.773 |
| | | | | | revmonotone | | 0.006 | 0.001 | 1.038 | 0.993 | 0.742 |
| | | | | | revmonotone | 3 | 0.006 | 0.001 | 0.972 | 1.000 | 0.785 |
| | | | | | revmonotone | 5 | 0.007 | 0.001 | 0.968 | 1.000 | 0.775 |
| | | | F | F | monotone | | 0.006 | 0.002 | 1.039 | 0.979 | 0.736 |
| | | | | | monotone | 3 | 0.008 | 0.002 | 0.979 | 0.997 | 0.775 |
| | | | | | monotone | 5 | 0.009 | 0.002 | 0.980 | 0.997 | 0.766 |
| | | | | | revmonotone | | 0.006 | 0.002 | 1.038 | 0.980 | 0.736 |
| | | | | | revmonotone | 3 | 0.008 | 0.002 | 0.980 | 0.998 | 0.776 |
| | | | | | revmonotone | 5 | 0.008 | 0.002 | 0.979 | 0.998 | 0.767 |
| IPW | logreg | F | T | | | | 0.001 | 0.007 | 0.946 | 0.998 | 1.218 |
| | | | F | | | | 0.001 | 0.007 | 0.946 | 0.998 | 1.218 |
| | | T | T | | | | 0.001 | 0.007 | 0.950 | 0.998 | 1.218 |
| | | | F | | | | 0.001 | 0.007 | 0.950 | 0.998 | 1.218 |
| | RF | F | T | | | | 0.002 | 0.008 | 0.937 | 0.997 | 1.233 |
| | | | F | | | | 0.004 | 0.007 | 0.946 | 0.997 | 1.230 |
| | | T | T | | | | 0.013 | 0.012 | 0.835 | 0.959 | 1.229 |
| | | | F | | | | 0.017 | 0.012 | 0.857 | 0.959 | 1.228 |

Figure 8: Average measures obtained by the tested imputation algorithms across the three outcomes (the table is also made available in Supplementary file S1 – sheet "mean_measures_m42") when MAR missingness is simulated in the amputated datasets. For the RB and MSE measures the highlighted cells mark the models that had less losses according to the paired Wilcoxon rank-sum tests computed over the three outcomes. For the CR measure, all the models, but the (non-augmented) IPW model (where the probability of data being missing was computed by an RF including the outcome variables in the model) had comparable performance. missRanger with no pmm achieves the lowest standard error estimate (indeed the ratio SE measures - column "ratio SE" - is the lowest, as also confirmed by the paired Wilkocox rank-sign test), IPW models obtain a standard error greater than the one computed on the unweighted dataset.

| MI algorithm | univariate imputation method | use outcomes | one-hot encode binned numeric predictors | one-hot encode categorical predictors | univariate imputation order | pmm donors | average ‖RB‖ across outcomes | average MSE across outcomes | average ER across outcomes | average CR across outcomes | average ratio_SE across outcomes |
|---|---|---|---|---|---|---|---|---|---|---|---|
| amelia | | F | T | | | | 0.025 | 0.007 | 0.897 | 0.937 | 0.722 |
| | | F | F | T | | | 0.017 | 0.003 | 0.945 | 0.972 | 0.734 |
| | | T | T | | | | 0.020 | 0.005 | 0.913 | 0.968 | 0.724 |
| | | T | F | | | | 0.014 | 0.003 | 0.973 | 0.988 | 0.735 |
| mice | logreg | F | T | T | monotone | | 0.013 | 0.002 | 0.951 | 0.978 | 0.736 |
| | | F | | | revmonotone | | 0.013 | 0.002 | 0.951 | 0.978 | 0.736 |
| | | T | | | monotone | | 0.011 | 0.002 | 0.955 | 0.987 | 0.737 |
| | | T | | | revmonotone | | 0.011 | 0.002 | 0.953 | 0.987 | 0.736 |
| | norm | F | T | T | monotone | | 0.012 | 0.002 | 0.954 | 0.978 | 0.737 |
| | | F | | | revmonotone | | 0.012 | 0.002 | 0.954 | 0.977 | 0.737 |
| | | F | F | | monotone | | 0.012 | 0.002 | 0.966 | 0.981 | 0.738 |
| | | F | F | | revmonotone | | 0.012 | 0.002 | 0.965 | 0.981 | 0.738 |
| | | T | T | | monotone | | 0.010 | 0.002 | 0.955 | 0.986 | 0.737 |
| | | T | | | revmonotone | | 0.010 | 0.002 | 0.954 | 0.986 | 0.737 |
| | | T | F | | monotone | | 0.010 | 0.002 | 0.969 | 0.989 | 0.738 |
| | | T | F | | revmonotone | | 0.010 | 0.002 | 0.968 | 0.989 | 0.738 |
| missRanger | extratrees | F | T | T | monotone | | 0.006 | 0.001 | 0.957 | 0.993 | 0.733 |
| | | | | | monotone | 3 | 0.007 | 0.001 | 0.962 | 0.995 | 0.764 |
| | | | | | monotone | 5 | 0.008 | 0.001 | 0.958 | 0.992 | 0.761 |
| | | | | | revmonotone | | 0.006 | 0.001 | 0.957 | 0.992 | 0.733 |
| | | | | | revmonotone | 3 | 0.007 | 0.001 | 0.964 | 0.993 | 0.764 |
| | | | | | revmonotone | 5 | 0.008 | 0.001 | 0.959 | 0.993 | 0.762 |
| | | F | F | F | monotone | | 0.005 | 0.001 | 0.966 | 0.996 | 0.730 |
| | | | | | monotone | 3 | 0.009 | 0.002 | 0.966 | 0.990 | 0.751 |
| | | | | | monotone | 5 | 0.009 | 0.002 | 0.968 | 0.987 | 0.746 |
| | | | | | revmonotone | | 0.005 | 0.001 | 0.967 | 0.996 | 0.730 |
| | | | | | revmonotone | 3 | 0.009 | 0.002 | 0.967 | 0.989 | 0.751 |
| | | | | | revmonotone | 5 | 0.009 | 0.002 | 0.966 | 0.986 | 0.747 |
| | | T | T | T | monotone | | 0.006 | 0.001 | 1.041 | 0.996 | 0.742 |
| | | | | | monotone | 3 | 0.005 | 0.001 | 0.977 | 1.000 | 0.785 |
| | | | | | monotone | 5 | 0.006 | 0.001 | 0.975 | 0.999 | 0.777 |
| | | | | | revmonotone | | 0.006 | 0.001 | 1.040 | 0.996 | 0.742 |
| | | | | | revmonotone | 3 | 0.006 | 0.001 | 0.975 | 1.000 | 0.789 |
| | | | | | revmonotone | 5 | 0.006 | 0.001 | 0.973 | 0.999 | 0.777 |
| | | T | F | F | monotone | | 0.006 | 0.001 | 1.050 | 0.979 | 0.736 |
| | | | | | monotone | 3 | 0.008 | 0.002 | 0.984 | 0.998 | 0.777 |
| | | | | | monotone | 5 | 0.008 | 0.002 | 0.983 | 0.998 | 0.768 |
| | | | | | revmonotone | | 0.006 | 0.001 | 1.050 | 0.978 | 0.736 |
| | | | | | revmonotone | 3 | 0.008 | 0.002 | 0.984 | 0.998 | 0.778 |
| | | | | | revmonotone | 5 | 0.008 | 0.002 | 0.986 | 0.999 | 0.768 |
| IPW | logreg | F | T | | | | 0.002 | 0.011 | 0.964 | 0.996 | 1.280 |
| | | F | F | | | | 0.002 | 0.011 | 0.965 | 0.996 | 1.280 |
| | | T | T | | | | 0.002 | 0.012 | 0.960 | 0.996 | 1.280 |
| | | T | F | | | | 0.002 | 0.011 | 0.961 | 0.996 | 1.280 |
| | RF | F | T | | | | 0.001 | 0.012 | 0.975 | 0.995 | 1.287 |
| | | F | F | | | | 0.001 | 0.012 | 0.965 | 0.994 | 1.286 |
| | | T | T | | | | 0.005 | 0.010 | 1.042 | 0.999 | 1.290 |
| | | T | F | | | | 0.007 | 0.010 | 1.026 | 0.998 | 1.291 |

Figure 9: Average measures obtained by the tested imputation algorithms across the three outcomes (the table is also made available in Supplementary file S3_MCAR – sheet "mean_measures_m42") when MCAR missingness is simulated in the amputated datasets.

# Supplementary files

## S1.xlsx:

This excel file contains all the results we obtained when using a number of imputation m = 42.
The file is composed by:

- sheet "mean_measures_m42" containing the coloured table (Figure 5) showing and detailing the average measures obtained by the tested imputation algorithms across the three outcomes.

- sheets "RB_mean" (see also Figure 6), "MSE_mean" (see also Figure 7), "ER_mean" (see also Figure 8), and "CR_mean" contain the four win-tie-loss tables (for the RB measure, the MSE measure, the ER measure, the CR measure) obtained by summing the wins, ties, losses obtained by each model over the three outcome variables.
On the right, each of the four sheets contains the mean of the win-tie-loss tables over the three outcomes, where the wins, ties, and losses are computed by comparing the models on the rows to the models on the column by a paired-sided paired rank sign test .
The grid shows numbers in the range [-3, +3]; they are computed by representing each win by a +1 value, each tie as a 0 value, each loss as a -1 value.

## S2.xlsx:

This excel file has the same structure of S1.xlsx; it details all the results we obtained when using a number of imputation m = 5.

## S3_MCAR.xlsx:

This excel file has the same structure of S1.xlsx and S2.xslx; it details all the results we obtained when simulating MCAR missingness in the amputated datasets.

## Supplementary Figures:

| MI algorithm | univariate imputation method | use outcomes | one-hot encode binned numeric predictors | one-hot encode categorical predictors | univariate imputation order | pmm donors | average of absolute values of RB across outcomes | wins | ties | losses |
|---|---|---|---|---|---|---|---|---|---|---|
| amelia | | F | T | T | | | 0.025 | 1 | 4 | -94 |
| | | F | F | | | | 0.017 | 7 | 16 | -50 |
| | | T | T | | | | 0.021 | 8 | 26 | -28 |
| | | T | F | | | | 0.014 | 44 | 21 | -10 |
| mice | default | F | F | F | monotone | | 0.012 | 6 | 16 | -63 |
| | | F | | | revmonotone | | 0.012 | 5 | 18 | -60 |
| | | T | | | monotone | | 0.012 | 7 | 14 | -63 |
| | | T | | | revmonotone | | 0.012 | 7 | 15 | -64 |
| | logreg | F | T | T | monotone | | 0.013 | 5 | 6 | -85 |
| | | F | | | revmonotone | | 0.013 | 5 | 3 | -89 |
| | | T | | | monotone | | 0.011 | 18 | 21 | -34 |
| | | T | | | revmonotone | | 0.011 | 22 | 18 | -31 |
| | norm | F | T | T | monotone | | 0.012 | 7 | 8 | -76 |
| | | F | | | revmonotone | | 0.012 | 6 | 5 | -80 |
| | | F | F | | monotone | | 0.012 | 10 | 11 | -63 |
| | | F | F | | revmonotone | | 0.013 | 9 | 10 | -66 |
| | | T | T | | monotone | | 0.011 | 34 | 16 | -23 |
| | | T | T | | revmonotone | | 0.011 | 35 | 17 | -23 |
| | | T | F | | monotone | | 0.011 | 40 | 18 | -16 |
| | | T | F | | revmonotone | | 0.011 | 41 | 17 | -15 |
| missRanger | extratrees | F | T | T | monotone | | 0.007 | 109 | 4 | 0 |
| | | | | | monotone | 3 | 0.008 | 40 | 8 | -38 |
| | | | | | monotone | 5 | 0.009 | 19 | 16 | -49 |
| | | | | | revmonotone | | 0.007 | 107 | 4 | 0 |
| | | | | | revmonotone | 3 | 0.008 | 51 | 8 | -32 |
| | | | | | revmonotone | 5 | 0.009 | 22 | 13 | -49 |
| | | F | F | F | monotone | | 0.007 | 113 | 4 | -2 |
| | | | | | monotone | 3 | 0.010 | 26 | 17 | -30 |
| | | | | | monotone | 5 | 0.010 | 12 | 20 | -38 |
| | | | | | revmonotone | | 0.007 | 114 | 4 | -2 |
| | | | | | revmonotone | 3 | 0.010 | 27 | 15 | -29 |
| | | | | | revmonotone | 5 | 0.010 | 10 | 18 | -45 |
| | | T | T | T | monotone | | 0.006 | 83 | 16 | 0 |
| | | | | | monotone | 3 | 0.006 | 81 | 10 | -9 |
| | | | | | monotone | 5 | 0.007 | 66 | 14 | -14 |
| | | | | | revmonotone | | 0.006 | 83 | 16 | 0 |
| | | | | | revmonotone | 3 | 0.006 | 84 | 8 | -8 |
| | | | | | revmonotone | 5 | 0.007 | 59 | 14 | -15 |
| | | T | F | F | monotone | | 0.006 | 51 | 13 | -9 |
| | | | | | monotone | 3 | 0.008 | 47 | 18 | -12 |
| | | | | | monotone | 5 | 0.009 | 36 | 18 | -21 |
| | | | | | revmonotone | | 0.006 | 50 | 14 | -8 |
| | | | | | revmonotone | 3 | 0.008 | 46 | 18 | -13 |
| | | | | | revmonotone | 5 | 0.008 | 34 | 18 | -25 |
| IPW | logreg | F | T | F | | | 0.001 | 31 | 24 | -6 |
| | | F | F | T | | | 0.001 | 32 | 24 | -6 |
| | | T | T | F | | | 0.001 | 28 | 27 | -6 |
| | | T | F | T | | | 0.001 | 28 | 27 | -6 |
| | RF | F | T | F | | | 0.002 | 5 | 11 | -48 |
| | | F | F | T | | | 0.004 | 3 | 4 | -60 |
| | | T | T | F | | | 0.013 | 0 | 3 | -79 |
| | | T | F | T | | | 0.017 | 0 | 2 | -82 |

Figure S1: column "average absolute value of RB across outcomes" reports the average RB measure across the hospitalization, invasive ventilation, and patients' survival outcomes (the table is also made available in Supplementary file S1 – sheet "RB_mean"). Columns "wins", "ties", "losses" report the sum of, respectively, wins, ties, and losses computed by comparing the (absolute value of the) RB measures over the three outcomes (the corresponding win-tie-loss grid is shown in the Supplementary material). The comparison between two models over an outcome variable is performed with a sided Wilcoxon signed-rank test comparing the distribution of the (absolute) RB values for all the predictor variables. The winner is the model achieving the lowest RB distribution. All the models but missRanger with no pmm and using the outcome variables in the imputation model are obtaining RB ≤ 0, meaning that the computed estimates are systematically lower than those computed on the complete dataset. missRanger with outcome variable in the imputation model and no pmm is instead bringing to the computation of inflated estimates.

| MI algorithm | univariate imputation method | use outcomes | one-hot encode binned numeric predictors | one-hot encode categorical predictors | univariate imputation order | pmm donors | average MSE across outcomes | wins | ties | losses |
|---|---|---|---|---|---|---|---|---|---|---|
| amelia | | | T | T | | | 0.007 | 18 | 5 | -76 |
| | | F | F | | | | 0.004 | 25 | 28 | -22 |
| | | | T | | | | 0.005 | 26 | 23 | -22 |
| | | T | F | | | | 0.003 | 66 | 19 | -1 |
| mice | default | F | | F | monotone | | 0.002 | 26 | 21 | -32 |
| | | | | | revmonotone | | 0.002 | 25 | 20 | -34 |
| | | T | | | monotone | | 0.002 | 30 | 18 | -32 |
| | | | | | revmonotone | | 0.002 | 26 | 19 | -34 |
| | logreg | F | | T | monotone | | 0.003 | 25 | 9 | -58 |
| | | | | | revmonotone | | 0.003 | 25 | 7 | -60 |
| | | T | | | monotone | | 0.002 | 37 | 26 | -15 |
| | | | | | revmonotone | | 0.002 | 39 | 25 | -13 |
| | norm | F | T | | monotone | | 0.002 | 28 | 13 | -43 |
| | | | | | revmonotone | | 0.002 | 27 | 9 | -48 |
| | | | F | | monotone | | 0.002 | 30 | 16 | -33 |
| | | | | | revmonotone | | 0.002 | 28 | 17 | -34 |
| | | T | T | | monotone | | 0.002 | 54 | 15 | -11 |
| | | | | | revmonotone | | 0.002 | 53 | 16 | -9 |
| | | | F | | monotone | | 0.002 | 67 | 12 | -5 |
| | | | | | revmonotone | | 0.002 | 72 | 14 | -2 |
| missRanger | extratrees | | T | T | monotone | | 0.001 | 93 | 2 | 0 |
| | | | | | monotone | 3 | 0.001 | 38 | 22 | -22 |
| | | | | | monotone | 5 | 0.001 | 28 | 21 | -34 |
| | | | | | revmonotone | | 0.001 | 92 | 3 | 0 |
| | | | | | revmonotone | 3 | 0.001 | 48 | 20 | -18 |
| | | | | | revmonotone | 5 | 0.001 | 31 | 21 | -29 |
| | | F | | | monotone | | 0.001 | 88 | 6 | -2 |
| | | | | | monotone | 3 | 0.002 | 34 | 27 | -13 |
| | | | | | monotone | 5 | 0.002 | 30 | 22 | -20 |
| | | | F | F | revmonotone | | 0.001 | 89 | 6 | -2 |
| | | | | | revmonotone | 3 | 0.002 | 43 | 25 | -8 |
| | | | | | revmonotone | 5 | 0.002 | 30 | 23 | -24 |
| | | T | T | T | monotone | | 0.001 | 47 | 20 | -2 |
| | | | | | monotone | 3 | 0.001 | 70 | 12 | -11 |
| | | | | | monotone | 5 | 0.001 | 84 | 13 | -12 |
| | | | | | revmonotone | | 0.001 | 46 | 21 | -2 |
| | | | | | revmonotone | 3 | 0.001 | 74 | 8 | -8 |
| | | | | | revmonotone | 5 | 0.001 | 85 | 10 | -17 |
| | | | F | F | monotone | | 0.002 | 36 | 7 | -28 |
| | | | | | monotone | 3 | 0.002 | 47 | 15 | -14 |
| | | | | | monotone | 5 | 0.002 | 42 | 16 | -18 |
| | | | | | revmonotone | | 0.002 | 36 | 5 | -30 |
| | | | | | revmonotone | 3 | 0.002 | 44 | 17 | -15 |
| | | | | | revmonotone | 5 | 0.002 | 37 | 18 | -20 |
| IPW | logreg | F | T | F | | | 0.007 | 13 | 3 | -130 |
| | | | F | | | | 0.007 | 14 | 2 | -130 |
| | | T | T | | | | 0.007 | 12 | 2 | -132 |
| | | | F | | | | 0.007 | 12 | 3 | -131 |
| | RF | F | T | F | | | 0.008 | 6 | 1 | -144 |
| | | | F | | | | 0.007 | 7 | 1 | -143 |
| | | T | T | | | | 0.012 | 1 | 1 | -150 |
| | | | F | | | | 0.012 | 0 | 1 | -151 |

Figure S2: column "average MSE across outcomes" reports the average MSE measure across the hospitalization, invasive ventilation, and patients' survival outcomes (the table is also made available in Supplementary file S1 – sheet "MSE_mean"). Columns "wins", "ties", "losses" report the sum of, respectively, wins, ties, and losses computed by comparing the MSE measures over the three outcomes (the corresponding win-tie-loss grid is shown in the Supplementary material). The comparison between two models over an outcome variable is performed with a sided Wilcoxon signed-rank test comparing the distribution of the MSE values for all the predictor variables. The winner is the model achieving the lowest MSE distribution. The detailed win-tie-loss grids are reported in the Supplementary material.

| MI algorithm | univariate imputation method | use outcomes | one-hot encode binned numeric predictors | one-hot encode categorical predictors | univariate imputation order | pmm donors | average ER across outcomes | wins | ties | losses |
|---|---|---|---|---|---|---|---|---|---|---|
| amelia | | F | T | T | | | 0.900 | 0 | 5 | -84 |
| amelia | | F | F | T | | | 0.949 | 4 | 21 | -43 |
| amelia | | T | T | T | | | 0.914 | 6 | 31 | -25 |
| amelia | | T | F | T | | | 0.971 | 34 | 23 | -13 |
| mice | default | F | F | F | monotone | | 0.962 | 6 | 14 | -54 |
| mice | default | F | F | F | revmonotone | | 0.964 | 6 | 14 | -52 |
| mice | default | T | F | F | monotone | | 0.963 | 8 | 17 | -47 |
| mice | default | T | F | F | revmonotone | | 0.962 | 9 | 12 | -53 |
| mice | logreg | F | T | T | monotone | | 0.950 | 5 | 10 | -76 |
| mice | logreg | F | T | T | revmonotone | | 0.950 | 5 | 7 | -77 |
| mice | logreg | T | T | T | monotone | | 0.952 | 16 | 21 | -31 |
| mice | logreg | T | T | T | revmonotone | | 0.952 | 17 | 25 | -30 |
| mice | norm | F | T | T | monotone | | 0.952 | 9 | 13 | -67 |
| mice | norm | F | T | T | revmonotone | | 0.950 | 7 | 12 | -68 |
| mice | norm | F | F | T | monotone | | 0.963 | 7 | 16 | -53 |
| mice | norm | F | F | T | revmonotone | | 0.961 | 6 | 14 | -55 |
| mice | norm | T | T | T | monotone | | 0.952 | 30 | 20 | -20 |
| mice | norm | T | T | T | revmonotone | | 0.951 | 28 | 22 | -20 |
| mice | norm | T | F | T | monotone | | 0.964 | 37 | 22 | -15 |
| mice | norm | T | F | T | revmonotone | | 0.963 | 39 | 21 | -15 |
| missRanger | extratrees | F | T | T | monotone | | 0.952 | 110 | 6 | 0 |
| missRanger | extratrees | F | T | T | monotone | 3 | 0.958 | 44 | 15 | -23 |
| missRanger | extratrees | F | T | T | monotone | 5 | 0.956 | 29 | 19 | -33 |
| missRanger | extratrees | F | T | T | revmonotone | | 0.951 | 111 | 6 | 0 |
| missRanger | extratrees | F | T | T | revmonotone | 3 | 0.958 | 54 | 18 | -17 |
| missRanger | extratrees | F | T | T | revmonotone | 5 | 0.955 | 30 | 20 | -29 |
| missRanger | extratrees | F | F | F | monotone | | 0.955 | 103 | 6 | 0 |
| missRanger | extratrees | F | F | F | monotone | 3 | 0.960 | 24 | 21 | -19 |
| missRanger | extratrees | F | F | F | monotone | 5 | 0.961 | 13 | 24 | -24 |
| missRanger | extratrees | F | F | F | revmonotone | | 0.956 | 103 | 6 | 0 |
| missRanger | extratrees | F | F | F | revmonotone | 3 | 0.959 | 25 | 21 | -19 |
| missRanger | extratrees | F | F | F | revmonotone | 5 | 0.960 | 11 | 26 | -29 |
| missRanger | extratrees | T | T | T | monotone | | 1.037 | 45 | 20 | 0 |
| missRanger | extratrees | T | T | T | monotone | 3 | 0.971 | 79 | 9 | -7 |
| missRanger | extratrees | T | T | T | monotone | 5 | 0.967 | 85 | 12 | -14 |
| missRanger | extratrees | T | T | T | revmonotone | | 1.038 | 42 | 22 | 0 |
| missRanger | extratrees | T | T | T | revmonotone | 3 | 0.972 | 82 | 9 | -6 |
| missRanger | extratrees | T | T | T | revmonotone | 5 | 0.968 | 60 | 13 | -16 |
| missRanger | extratrees | T | F | F | monotone | | 1.039 | 12 | 27 | -15 |
| missRanger | extratrees | T | F | F | monotone | 3 | 0.979 | 29 | 24 | -10 |
| missRanger | extratrees | T | F | F | monotone | 5 | 0.980 | 24 | 21 | -15 |
| missRanger | extratrees | T | F | F | revmonotone | | 1.038 | 14 | 25 | -15 |
| missRanger | extratrees | T | F | F | revmonotone | 3 | 0.980 | 31 | 21 | -12 |
| missRanger | extratrees | T | F | F | revmonotone | 5 | 0.979 | 24 | 21 | -19 |
| IPW | logreg | F | T | F | | | 0.946 | 18 | 35 | -6 |
| IPW | logreg | F | F | T | | | 0.946 | 19 | 34 | -6 |
| IPW | logreg | T | T | F | | | 0.950 | 17 | 36 | -6 |
| IPW | logreg | T | F | T | | | 0.950 | 19 | 34 | -6 |
| IPW | RF | F | T | F | | | 0.937 | 4 | 8 | -52 |
| IPW | RF | F | F | T | | | 0.946 | 3 | 5 | -54 |
| IPW | RF | T | T | F | | | 0.835 | 0 | 3 | -84 |
| IPW | RF | T | F | T | | | 0.857 | 0 | 3 | -89 |

Figure S3: In column "average ER across outcomes" we report the average ER measure across the hospitalization, invasive ventilation, and patients' survival outcomes (the table is also made available in Supplementary file S1 – sheet "ER_mean"). Columns "wins", "ties", "losses" report the sum of, respectively, wins, ties, and losses computed by comparing the distributions of the ER measures over the three outcomes. Since we would like each $[ER_i]$ ($i \in \{1, ..., d\}$) estimate to be as nearest as possible to 1, for the comparison between two models over an outcome variable we used a sided Wilcoxon signed-rank test to compare the following distribution for each model $f(ER_i) = \|1 - ER_i\|$. The detailed win tie loss grids are reported in the supplementary material.

# Appendix A

## *Defining the number **m** of multiple imputations*

The minimum number of imputation is chosen by considering the statistical efficiency of the estimates obtained after imputation [4][37]. The relative efficiency *RE* of an estimator (e.g. logistic regression model or cox-survival model) exploiting **m** multiple imputations may be evaluated as the ratio of the total variance obtained with m imputations, $T_m$, compared to the total variance when $m \to \infty$, $T_\infty$. Rubin [4] showed that this ratio is:

$$RE = \frac{T_m}{T_\infty} = 1 + \frac{1}{m}\gamma_0$$

where $\gamma_0$ is the Fraction of Missing Information FMI = B/(W + B), with W representing within imputation variance and B representing between imputation variance [14]. From the above equation we get the percent loss of efficiency which should be minimized:

$$\frac{T_m}{T_\infty} = 1 + \frac{1}{m}\gamma_0 \Rightarrow \frac{T_m - T_\infty}{T_\infty} = \frac{1}{m}\gamma_0 = \frac{1}{m}FMI$$

Therefore, the value of m should be chosen so that $\frac{1}{m}FMI$ becomes negligible or as small as it is feasible. White et al. [14] suggest using, for example, $\frac{1}{m}FMI < 0.05$, that is, they require a maximum percent loss in efficiency lower than the 5 percent. Of course, this computation requires estimating the FMI due to the missing values, which on the other side would require trying different values for m, computing the pooled estimates and their FMI. A faster way is to use an upper bound of the FMI, that White et al. [14] conservatively estimated as the fraction of incomplete cases. This led to Von Hippel's rule of thumb [40] that *the number of imputations should be similar to the percentage of cases that are incomplete,* which means we are requiring the maximum percent loss in efficiency lower than 1 percent. Unfortunately, when the sample size is high, such a low maximum percent loss in efficiency would result in demanding, and often impractical computational costs. Besides, as noted also in [14] this estimate may not be appropriate, because it also depends on the data and the problem at hand. Indeed, several other state-of-the-art definitions of FMI and different experimental results proposing estimates for the number of multiple imputations have been shown in literature.

As an example, Graham [38] studied the loss in power when small numbers of imputed datasets are used. They recommended that at least 20 imputed datasets are needed to restrict the loss of power when testing a relationship between variables. Bodner [39] proposed the following guidelines after a simulation study using different values for the FMI to determine the number of imputed datasets. For FMI´s of 0.05, 0.1, 0.2, 0.3, 0.5 the following number of imputed datasets are needed: ≥3, 6, 12, 24, 59, respectively.

We refer interested readers to Van Buuren's book [28] for a brief description of different theories.

## Authorship

Authorship was determined using ICMJE recommendations.

## Authors' CRediT Contribution

**Elena Casiraghi**: Conceptualization; Formal analysis; Investigation; Methodology; Software; Supervision; Validation; Visualization; Writing - original draft; Writing - review & editing. **Rachel Wong:** Data curation; Resources; Validation; Writing - original draft; Writing - review & editing. **Margaret Hall:** Data curation; Software. **Ben Coleman:** Software; Writing - original draft; Writing - review & editing. **Marco Notaro:** Methodology; Software; Writing - original draft. **Michael D. Evans:** Software; Writing - review & editing. **Jena S. Tronieri:** Validation; Writing - original draft; Writing - review & editing. **Hannah Blau:** Validation; Writing - original draft; Writing - review & editing. **Bryan Laraway:** Data curation. **Tiffany J. Callahan:** Data curation. **Lauren E. Chan:** Validation; Writing - review & editing. **Carolyn T. Bramante:** Writing - review & editing. **John B. Buse:** Supervision; Writing - review & editing. **Richard A. Moffitt:** Data curation. **Til Sturmer:** Validation; Writing - review & editing. **Steven G. Johnson:** Data curation. **Yu Raymond Shao:** Validation; Writing - original draft; Writing - review & editing. **Justin Reese:** Methodology; Validation; Writing - review & editing. **Peter N. Robinson:** Methodology; Validation; Writing - review & editing. **Alberto Paccanaro:** Methodology; Funding acquisition; Supervision; Writing - original draft; Writing - review & editing. **Giorgio Valentini:** Methodology; Funding acquisition; Supervision; Writing - original draft; Writing - review & editing. **Jared D. Huling:** Methodology; Validation; Supervision; Writing - original draft; Writing - review & editing. **Kenneth J. Wilkins:** Methodology; Validation; Supervision; Writing - original draft; Writing - review & editing.

For transparency, we encourage authors to submit an author statement file outlining their individual contributions to the paper using the relevant CRediT roles: Conceptualization; Data curation; Formal analysis; Funding acquisition; Investigation; Methodology; Project administration; Resources; Software; Supervision; Validation; Visualization; Roles/Writing - original draft; Writing - review & editing. Authorship statements should be formatted with the names of authors first and CRediT role(s) following. More details and an example.

## Consent

All the authors read the paper and gave their consent for submission.

**Conflict of interest**

All the authors declare no conflict of interest

**IRB and DUR**

The N3C data transfer to NCATS is performed under a Johns Hopkins University Reliance Protocol # IRB00249128 or individual site agreements with NIH. The N3C Data Enclave is managed under the authority of the NIH; information can be found at https://ncats.nih.gov/n3c/resources.
The study was conducted under the DUR RP-60B81D

**N3C Attribution**

The analyses described in this publication were conducted with data or tools accessed through the NCATS N3C Data Enclave covid.cd2h.org/enclave and supported by CD2H - The National COVID Cohort Collaborative (N3C) IDeA CTR Collaboration 3U24TR002306-04S2 NCATS U24 TR002306. This research was possible because of the patients whose information is included within the data from participating organizations (covid.cd2h.org/dtas) and the organizations and scientists (covid.cd2h.org/duas) who have contributed to the on-going development of this community resource (cite this https://doi.org/10.1093/jamia/ocaa196).

**Disclaimer**

The content is solely the responsibility of the authors and does not necessarily represent the official views of the National Institutes of Health or the N3C program.

**Fundings**

Elena Casiraghi, Marco Notaro, and Giorgio Valentini were supported by Università degli Studi di Milano, Piano di sviluppo di ricerca, grant number 2015-17 PSR2015-17

Alberto Paccanaro was supported by Biotechnology and Biological Sciences Research Council (https://bbsrc.ukri.org/) grants numbers BB/K004131/1, BB/F00964X/1 and BB/M025047/1, Medical Research Council

(https://mrc.ukri.org) grant number MR/T001070/1, Consejo Nacional de Ciencia y Tecnología Paraguay (https://www.conacyt.gov.py/) grants numbers 14-INV-088, PINV15–315 and PINV20-337, National Science Foundation Advances in Bio Informatics (https://www.nsf.gov/) grant number 1660648, Fundação de Amparo à Pesquisa do Estado do Rio de Janeiro grant number E-26/201.079/2021 (260380) and Fundação Getulio Vargas.

**Acknowledgements**

The analyses described in this publication were conducted with data or tools accessed through the NCATS N3C Data Enclave covid.cd2h.org/enclave and supported by NCATS U24 TR002306. This research was possible because of the patients whose information is included within the data from participating organizations (covid.cd2h.org/dtas) and the organizations and scientists (covid.cd2h.org/duas) who have contributed to the on-going development of this community resource (cite this https://doi.org/10.1093/jamia/ocaa196).

**Individual Acknowledgements For Core Contributors**

We gratefully acknowledge the following core contributors to N3C:
Adam B. Wilcox, Adam M. Lee, Alexis Graves, Alfred (Jerrod) Anzalone, Amin Manna, Amit Saha, Amy Olex, Andrea Zhou, Andrew E. Williams, Andrew Southerland, Andrew T. Girvin, Anita Walden, Anjali A. Sharathkumar, Benjamin Amor, Benjamin Bates, Brian Hendricks, Brijesh Patel, Caleb Alexander, Carolyn Bramante, Cavin Ward-Caviness, Charisse Madlock-Brown, Christine Suver, Christopher Chute, Christopher Dillon, Chunlei Wu, Clare Schmitt, Cliff Takemoto, Dan Housman, Davera Gabriel, David A. Eichmann, Diego Mazzotti, Don Brown, Eilis Boudreau, Elaine Hill, Elizabeth Zampino, Emily Carlson Marti, Emily R. Pfaff, Evan French, Farrukh M Koraishy, Federico Mariona, Fred Prior, George Sokos, Greg Martin, Harold Lehmann, Heidi Spratt, Hemalkumar Mehta, Hongfang Liu, Hythem Sidky, J.W. Awori Hayanga, Jami Pincavitch, Jaylyn Clark, Jeremy Richard Harper, Jessica Islam, Jin Ge, Joel Gagnier, Joel H. Saltz, Joel Saltz, Johanna Loomba, John Buse, Jomol Mathew, Joni L. Rutter, Julie A. McMurry, Justin Guinney, Justin Starren, Karen Crowley, Katie Rebecca Bradwell, Kellie M. Walters, Ken Wilkins, Kenneth R. Gersing, Kenrick Dwain Cato, Kimberly Murray, Kristin Kostka, Lavance Northington, Lee Allan Pyles, Leonie Misquitta, Lesley Cottrell, Lili Portilla, Mariam Deacy, Mark M. Bissell, Marshall Clark, Mary Emmett, Mary Morrison Saltz, Matvey B. Palchuk, Melissa A. Haendel, Meredith Adams, Meredith Temple-O'Connor, Michael G. Kurilla, Michele Morris, Nabeel Qureshi, Nasia Safdar, Nicole Garbarini, Noha Sharafeldin, Ofer Sadan, Patricia A. Francis, Penny Wung Burgoon, Peter Robinson, Philip R.O. Payne, Rafael Fuentes, Randeep Jawa, Rebecca Erwin-Cohen, Rena Patel, Richard A. Moffitt, Richard L. Zhu, Rishi Kamaleswaran, Robert Hurley, Robert T. Miller, Saiju Pyarajan, Sam G. Michael, Samuel Bozzette, Sandeep Mallipattu, Satyanarayana Vedula, Scott Chapman, Shawn T. O'Neil,

**Data Partners with Released Data**

The following institutions whose data is released or pending:

New Jersey Alliance for Clinical and Translational Science • Stony Brook University — U24TR002306 • The Ohio State University — UL1TR002733: Center for Clinical and Translational Science • The State University of New York at Buffalo — UL1TR001412: Clinical and Translational Science Institute • The University of Chicago — UL1TR002389: The Institute for Translational Medicine (ITM) • The University of Iowa — UL1TR002537: Institute for Clinical and Translational Science • The University of Miami Leonard M. Miller School of Medicine — UL1TR002736: University of Miami Clinical and Translational Science Institute • The University of Michigan at Ann Arbor — UL1TR002240: Michigan Institute for Clinical and Health Research • The University of Texas Health Science Center at Houston — UL1TR003167: Center for Clinical and Translational Sciences (CCTS) • The University of Texas Medical Branch at Galveston — UL1TR001439: The Institute for Translational Sciences • The University of Utah — UL1TR002538: Uhealth Center for Clinical and Translational Science • Tufts Medical Center — UL1TR002544: Tufts Clinical and Translational Science Institute • Tulane University — UL1TR003096: Center for Clinical and Translational Science • University Medical Center New Orleans — U54GM104940: Louisiana Clinical and Translational Science (LA CaTS) Center • University of Alabama at Birmingham — UL1TR003096: Center for Clinical and Translational Science • University of Arkansas for Medical Sciences — UL1TR003107: UAMS Translational Research Institute • University of Cincinnati — UL1TR001425: Center for Clinical and Translational Science and Training • University of Colorado Denver, Anschutz Medical Campus — UL1TR002535: Colorado Clinical and Translational Sciences Institute • University of Illinois at Chicago — UL1TR002003: UIC Center for Clinical and Translational Science • University of Kansas Medical Center — UL1TR002366: Frontiers: University of Kansas Clinical and Translational Science Institute • University of Kentucky — UL1TR001998: UK Center for Clinical and Translational Science • University of Massachusetts Medical School Worcester — UL1TR001453: The UMass Center for Clinical and Translational Science (UMCCTS) • University of Minnesota — UL1TR002494: Clinical and Translational Science Institute • University of Mississippi Medical Center — U54GM115428: Mississippi Center for Clinical and Translational Research (CCTR) • University of Nebraska Medical Center — U54GM115458: Great Plains IDeA-Clinical & Translational Research • University of North Carolina at Chapel Hill — UL1TR002489: North Carolina Translational and Clinical Science Institute • University of Oklahoma Health Sciences Center — U54GM104938: Oklahoma Clinical and Translational Science Institute (OCTSI) • University of Rochester — UL1TR002001: UR Clinical & Translational Science Institute • University of Southern California — UL1TR001855: The Southern California Clinical and Translational Science Institute (SC CTSI) • University of Vermont — U54GM115516: Northern New England Clinical & Translational Research (NNE-CTR) Network • University of Virginia — UL1TR003015: iTHRIV Integrated Translational health Research Institute of Virginia • University of Washington — UL1TR002319: Institute of Translational Health Sciences • University of Wisconsin-Madison — UL1TR002373: UW Institute for Clinical and Translational Research • Vanderbilt University Medical Center — UL1TR002243: Vanderbilt Institute for Clinical and Translational Research • Virginia Commonwealth University — UL1TR002649: C.

Kenneth and Dianne Wright Center for Clinical and Translational Research • Wake Forest University Health Sciences — UL1TR001420: Wake Forest Clinical and Translational Science Institute • Washington University in St. Louis — UL1TR002345: Institute of Clinical and Translational Sciences • Weill Medical College of Cornell University — UL1TR002384: Weill Cornell Medicine Clinical and Translational Science Center • West Virginia University — U54GM104942: West Virginia Clinical and Translational Science Institute (WVCTSI)

Submitted: Icahn School of Medicine at Mount Sinai — UL1TR001433: ConduITS Institute for Translational Sciences • The University of Texas Health Science Center at Tyler — UL1TR003167: Center for Clinical and Translational Sciences (CCTS) • University of California, Davis — UL1TR001860: UCDavis Health Clinical and Translational Science Center • University of California, Irvine — UL1TR001414: The UC Irvine Institute for Clinical and Translational Science (ICTS) • University of California, Los Angeles — UL1TR001881: UCLA Clinical Translational Science Institute • University of California, San Diego — UL1TR001442: Altman Clinical and Translational Research Institute • University of California, San Francisco — UL1TR001872: UCSF Clinical and Translational Science Institute

Pending: Arkansas Children's Hospital — UL1TR003107: UAMS Translational Research Institute • Baylor College of Medicine — None (Voluntary) • Children's Hospital of Philadelphia — UL1TR001878: Institute for Translational Medicine and Therapeutics • Cincinnati Children's Hospital Medical Center — UL1TR001425: Center for Clinical and Translational Science and Training • Emory University — UL1TR002378: Georgia Clinical and Translational Science Alliance • HonorHealth — None (Voluntary) • Loyola University Chicago — UL1TR002389: The Institute for Translational Medicine (ITM) • Medical College of Wisconsin — UL1TR001436: Clinical and Translational Science Institute of Southeast Wisconsin • MedStar Health Research Institute — UL1TR001409: The Georgetown-Howard Universities Center for Clinical and Translational Science (GHUCCTS) • MetroHealth — None (Voluntary) • Montana State University — U54GM115371: American Indian/Alaska Native CTR • NYU Langone Medical Center — UL1TR001445: Langone Health's Clinical and Translational Science Institute • Ochsner Medical Center — U54GM104940: Louisiana Clinical and Translational Science (LA CaTS) Center • Regenstrief Institute — UL1TR002529: Indiana Clinical and Translational Science Institute • Sanford Research — None (Voluntary) • Stanford University — UL1TR003142: Spectrum: The Stanford Center for Clinical and Translational Research and Education • The Rockefeller University — UL1TR001866: Center for Clinical and Translational Science • The Scripps Research Institute — UL1TR002550: Scripps Research Translational Institute • University of Florida — UL1TR001427: UF Clinical and Translational Science Institute • University of New Mexico Health Sciences Center — UL1TR001449: University of New Mexico Clinical and Translational Science Center • University of Texas Health Science Center at San Antonio — UL1TR002645: Institute for Integration of Medicine and Science • Yale New Haven Hospital — UL1TR001863: Yale Center for Clinical Investigation

## References

[1] Madden, J. M., Lakoma, M. D., Rusinak, D., Lu, C. Y., & Soumerai, S. B. (2016). Missing clinical and behavioral health data in a large electronic health record (EHR) system. Journal of the American Medical Informatics Association, 23(6), 1143-1149.

[2] Groenwold, R. H. (2020). Informative missingness in electronic health record systems: the curse of knowing. Diagnostic and prognostic research, 4(1), 1-6.

[3] Haneuse, S., Arterburn, D., & Daniels, M. J. (2021). Assessing missing data assumptions in EHR-based studies: a complex and underappreciated task. JAMA Network Open, 4(2), e210184-e210184.

[4] Rubin, D.B. (1987). Multiple Imputation for Nonresponse in Surveys. New York: John Wiley & Sons.

[5] Carlin, J.B. (2014). Multiple Imputation: Perspective and Historical Overview. Chapter 12 of Handbook of Missing Data Methodology, Edited by Molenberghs, G., Fitzmaurice, G. M., Kenward, M. G., Tsiatis, A., Verbeke, G. New York: Chapman & Hall/CRC. https://doi.org/10.1201/b17622

[6] Kenward, Michael G. and Carpenter, James R. (2009). Multiple Imputation. Chapter 21 of Longitudinal Data Analysis, Edited by Fitzmaurice, G. M., Davidian, M. Verbeke, G., Molenberghs, G. New York: Chapman & Hall/CRC. https://doi.org/10.1201/9781420011579

[7] Murray, J. S. (2018). Multiple imputation: a review of practical and theoretical findings. Statistical Science, 33(2018), 142-159.

[8] Cappelletti, L., Fontana, T., Di Donato, G. W., Di Tucci, L., Casiraghi, E., Valentini, G. (2020). Complex data imputation by auto-encoders and convolutional neural networks—A case study on genome gap-filling. Computers, 9(2), 37.

[9] van der Laan, Mark J. and Robins, James M. (2003). Unified Methods for Censored Longitudinal Data and Causality. New York: Springer.

[10] Zhang, Y., Alyass, A., Vanniyasingam, T., Sadeghirad, B., Flórez, I. D., Pichika, S. C., ... & Guyatt, G. H. (2017). A systematic survey of the methods literature on the reporting quality and optimal methods of handling participants with missing outcome data for continuous outcomes in randomized controlled trials. Journal of clinical epidemiology, 88, 67-80.

[11] Casiraghi, E., Malchiodi, D., Trucco, G., Frasca, M., Cappelletti, L., Fontana, T., ... & Valentini, G. (2020). Explainable machine learning for early assessment of COVID-19 risk prediction in emergency departments. Ieee Access, 8, 196299-196325.

[12] Hasan, M. K., Alam, M. A., Roy, S., Dutta, A., Jawad, M. T., & Das, S. (2021). Missing value imputation affects the performance of machine learning: A review and analysis of the literature (2010–2021). Informatics in Medicine Unlocked, 27, 100799.

[13] Moons, K.G., Donders, R.A., Stijnen, T., Harrell Jr, F.E. (2006). Using the outcome for imputation of missing predictor values was preferred. Journal of clinical epidemiology, 59(10), 1092-1101.

[14] White, I.R., Royston, P., Wood, A.M. (2011). Multiple imputation using chained equations: issues and guidance for practice. Statistics in medicine, 30(4), 377-399.

[15] Wong, R., Hall, M., Vaddavalli, R., Anand, A., Arora, N., Bramante, C. T., ... & N3C Consortium. (2022). Glycemic Control and Clinical Outcomes in US Patients With COVID-19: Data From the National COVID Cohort Collaborative (N3C) Database. Diabetes care, 45(5), 1099-1106.

[16] Seaman, S. R., White, I. R. (2013). Review of inverse probability weighting for dealing with missing data. Statistical methods in medical research, 22(3), 278-295. https://journals.sagepub.com/doi/10.1177/0962280210395740

[17] Fitzmaurice, Garrett M. (2014). Semiparametric Methods: Introduction and Overview. Chapter 7 of Handbook of Missing Data Methodology (2014), Edited by Molenberghs, G., Fitzmaurice, G. M., Kenward, M. G., Tsiatis, A., Verbeke, G. New York: Chapman & Hall/CRC. https://doi.org/10.1201/b17622

[18] Chan, L. E., Casiraghi, E., Laraway, B. J., … & Reese, J. (2022). Metformin is Associated with Reduced COVID-19 Severity in Patients with Prediabetes. medRxiv. https://www.medrxiv.org/content/10.1101/2022.08.29.22279355v1

[19] Goldstein, D. E., Little, R. R., Lorenz, R. A., Malone, J. I., Nathan, D., Peterson, C. M., & Sacks, D. B. (2004). Tests of glycemia in diabetes. Diabetes care, 27(7), 1761-1773.

[20] Anderson, M. R., Geleris, J., Anderson, D. R., Zucker, J., Nobel, Y. R., Freedberg, D., ... & Baldwin, M. R. (2020). Body mass index and risk for intubation or death in SARS-CoV-2 infection: a retrospective cohort study. Annals of internal medicine, 173(10), 782-790.

[21] Tartof, S. Y., Qian, L., Hong, V., Wei, R., Nadjafi, R. F., Fischer, H., ... & Murali, S. B. (2020). Obesity and mortality among patients diagnosed with COVID-19: results from an integrated health care organization. Annals of internal medicine, 173(10), 773-781.

[22] Sze, S., Pan, D., Nevill, C. R., Gray, L. J., Martin, C. A., Nazareth, J., ... & Pareek, M. (2020). Ethnicity and clinical outcomes in COVID-19: a systematic review and meta-analysis. EClinicalMedicine, 29, 100630.

[23] Magesh, S., John, D., Li, W. T., Li, Y., Mattingly-App, A., Jain, S., ... & Ongkeko, W. M. (2021). Disparities in COVID-19 outcomes by race, ethnicity, and socioeconomic status: a systematic-review and meta-analysis. JAMA network open, 4(11), e2134147-e2134147..

[24] CDC: https://www.cdc.gov/healthyweight/assessing/bmi/adult_bmi/index.html

[25] Weir, C. B. and Jan, A. (2019). BMI classification percentile and cut off points. In: StatPearls [Internet]. Treasure Island (FL): StatPearls Publishing; url: https://www.ncbi.nlm.nih.gov/books/NBK541070/

[26] Cook, L., Espinoza, J., Weiskopf, N. G., Mathews, N., Dorr, D. A., Gonzales, K. L., Wilcox, A., Madlock-Brown, C., & N3C Consortium (2022). Issues With Variability in Electronic Health Record Data About Race and Ethnicity: Descriptive Analysis of the National COVID Cohort Collaborative Data Enclave. JMIR medical informatics, 10(9), e39235. https://doi.org/10.2196/39235

[27] Li, C. (2013). Little's test of missing completely at random. The Stata Journal, 13(4), 795-809.

[28] Van Buuren, S. Flexible imputation of missing data. CRC press; 2018 Jul 17. url: https://stefvanbuuren.name/fimd/

[29] Jakobsen, J. C., Gluud, C., Wetterslev, J., & Winkel, P. (2017). When and how should multiple imputation be used for handling missing data in randomised clinical trials–a practical guide with flowcharts. BMC medical research methodology, 17(1), 1-10.

[30] Bhaskaran, K., Smeeth, L. (2014). What is the difference between missing completely at random and missing at random? Int J Epidemiol, Aug;43(4),1336-9. doi: 10.1093/ije/dyu080.

[31] Schouten, R.M., Vink, G. (2021). The dance of the mechanisms: How observed information influences the validity of missingness assumptions. Sociological Methods & Research, 50(3), 1243-1258.

[32] Little, R. J., & Rubin, D. B. (2019). Statistical analysis with missing data (Vol. 793). John Wiley & Sons.

[33] Schafer, J. L., & Graham, J. W. (2002). Missing data: our view of the state of the art. Psychological methods, 7(2), 147-177.

[34] Gelman, A., Hill, J. (2006). Data analysis using regression and multilevel/hierarchical models. Cambridge university press.

[35] Molenberghs, G., Beunckens, C., Sotto C., Kenward, M. G. (2008). Every Missingness Not at Random Model Has a Missingness at Random Counterpart with Equal Fit. Journal of the Royal Statistical Society. Series B (Statistical Methodology), 70(2), 371-388.

[36] Schafer JL. Analysis of Incomplete Multivariate Data. Chapman & Hall: London, 1997

[37] Schafer, J. L., Olsen, M. K. (1998). Multiple Imputation for Multivariate Missing-Data problems: A data analyst's perspective. Multivariate behavioral research, 33(4), 545-571.

[38] Graham, J. W., Olchowski, A. E., Gilreath, T. D. (2007). How Many Imputations Are Really Needed? Some Practical Clarifications of Multiple Imputation Theory. Preventive Science, 8(3), 206–13.

[39] Bodner, T.E. (2008). What improves with increased missing data imputations? Structural Equation Modeling: A Multidisciplinary Journal, 15, 651-675.

[40] Von Hippel, P.T. (2009). How to Impute Interactions, Squares, and Other Transformed Variables. Sociological Methodology, 39(1), 265–91.

[41] Rotnitzky, Andrea and Vansteelandt, Stijn. (2014). Double-Robust Methods. Chapter 9 of Handbook of Missing Data Methodology (2014), Edited by Molenberghs, G., Fitzmaurice, G. M., Kenward, M. G., Tsiatis, A., Verbeke, G. New York: Chapman & Hall/CRC. https://doi.org/10.1201/b17622

[42] Stekhoven, D.J., Bühlmann P. (2012). MissForest—non-parametric missing value imputation for mixed-type data. Bioinformatics, 28(1), 112-8.

[43] Pereira, R. C., Santos, M. S., Rodrigues, P. P., Abreu, P. H., (2020). Reviewing autoencoders for missing data imputation: Technical trends, applications and outcomes. Journal of Artificial Intelligence Research, 14(69), 1255-85

[44] Gondara, L., Wang, K. (2018). Mida: Multiple imputation using denoising autoencoders. In Pacific-Asia conference on knowledge discovery and data mining, 260-272. Springer, Cham.

[45] Kim, J. C., Chung, K. (2020). Multi-Modal Stacked Denoising Autoencoder for Handling Missing Data in Healthcare Big Data. IEEE Access (8), 104933-104943. doi: 10.1109/ACCESS.2020.2997255

[46] Jabbar, A., Xi, L., Bourahla, O. (2021). A survey on generative adversarial networks: Variants, applications, and training. ACM Computing Surveys (CSUR), 54(8), 1-49

[47] Yoon, J., Jordon, J., & van der Schaar, M., (2018). Gain: Missing data imputation using generative adversarial nets, International conference on machine learning, 5689–5698. doi: 10.48550/ARXIV.1806.02920. https://arxiv.org/abs/1806.02920

[48] Cheng-Xian Li, S., Jiang, B., Marlin, B. (2019). Learning from Incomplete Data with Generative Adversarial Networks, arXiv, doi: 10.48550/ARXIV.1902.09599. https://arxiv.org/abs/1902.09599

[49] Yuan, Y. (2011). Multiple Imputation Using SAS Software. Journal of Statistical Software, 1-25. http://www.jstatsoft.org/v45/i06/

[50] Honaker, J., King, G., Blackwell, M. (2011). "Amelia II: A Program for Missing Data." Journal of Statistical Software, 45(7), 1–47. https://www.jstatsoft.org/v45/i07/.

[51] Horton, N. J., Lipsitz, S. R., and Parzen, M., (2003). A potential for bias when rounding in multiple imputation, The American Statistician, 57 (4), 229–232

[52] Van Buuren, S. and Groothuis-Oudshoorn, K. (2011). mice: Multivariate imputation by chained equations. In R. Journal of statistical software, 45, 1-67.

[53] Breiman, L., Friedman, J. H., Olshen, R. A., Stone, C. J., (1984). Classification and regression trees. Belmont, CA: Wadsworh, Inc.

[54] Burgette, L., Reiter, J. P., (2010). Multiple imputation via sequential regression trees. American Journal of Epidemiology, 172, 1070–1076

[55] Akande, O., Li, F., Reiter, J., (2017). An empirical comparison of multiple imputation methods for categorical data. The American Statistician, 71 (2), 162-170

[56] Breiman, L. (2001). Random forests. Machine learning, 45(1), 5-32.

[57] Sportisse, A., Boyer, C., & Josse, J. (2020). Estimation and imputation in probabilistic principal component analysis with missing not at random data. Advances in Neural Information Processing Systems, 33, 7067-7077. https://proceedings.neurips.cc/paper/2020/file/4ecb679fd35dcfd0f0894c399590be1a-Paper.pdf

[58] Pereira, R. C., Abreu, P. H., & Rodrigues, P. P. (2022). Partial Multiple Imputation With Variational Autoencoders: Tackling Not at Randomness in Healthcare Data. IEEE Journal of Biomedical and Health Informatics, 26(8), 4218-4227. https://ieeexplore.ieee.org/document/9769986

[59] Schouten, R.M., Lugtig, P. and Vink, G. (2018). Generating missing values for simulation purposes: a multivariate amputation procedure. Journal of Statistical Computation and Simulation, 88(15), pp.2909-2930.

[60] Hong, S., Sun, Y., Li, H., Lynn, H.S (2021). A Note on the Required Sample Size of Model-Based Dose-Finding Methods for Molecularly Targeted Agents. Austin Biomedics and Biostatistics, 6(1), 1037.

[61] Haendel, M. A., Chute, C. G., Bennett, T. D., Eichmann, D. A., Guinney, J., Kibbe, W. A., ... & Gersing, K. R. (2021). The National COVID Cohort Collaborative (N3C): rationale, design, infrastructure, and deployment. Journal of the American Medical Informatics Association, 28(3), 427-443.

[62] Bennett, T. D., Moffitt, R. A., Hajagos, J. G., Amor, B., Anand, A., Bissell, M. M., ... & Koraishy, F. M. (2021). Clinical characterization and prediction of clinical severity of SARS-CoV-2 infection among US adults using data from the US National COVID Cohort Collaborative. JAMA network open, 4(7), e2116901-e2116901.

[63] Blake, M., DeWitt, P. E., Russell, S., Anand, A., Bradwell, K. R., Bremer, C., Gabriel, D., et al. 2021. "Children with SARS-CoV-2 in the National COVID Cohort Collaborative (N3C)." medRxiv : The Preprint Server for Health Sciences, July. https://doi.org/10.1101/2021.07.19.21260767.

[64] Sharafeldin, N., Bates, B., Song, Q., Madhira, V., Yan, Y., Dong, S., ... & Topaloglu, U. (2021). Outcomes of COVID-19 in patients with cancer: report from the National COVID Cohort Collaborative (N3C). Journal of Clinical Oncology, 39(20), 2232-2246.

[65] Bramante, C. T., Buse, J., Tamaritz, L., Palacio, A., Cohen, K., Vojta, D., ... & Tignanelli, C. J. (2021). Outpatient metformin use is associated with reduced severity of COVID-19 disease in adults with overweight or obesity. Journal of medical virology, 93(7), 4273-4279.

[66] Kahkoska, A. R., Abrahamsen, T. J., Alexander, G. C., Bennett, T. D., Chute, C. G., Haendel, M. A., ... & N3C Consortium Duong Tim Q. (2021). Association Between Glucagon-Like Peptide 1 Receptor Agonist and Sodium–Glucose Cotransporter 2 Inhibitor Use and COVID-19 Outcomes. Diabetes Care, 44(7), 1564-1572.

[67] Yang, X., Sun, J., Patel, R. C., Zhang, J., Guo, S., Zheng, Q., ... & Mannon, R. B. (2021). Associations between HIV infection and clinical spectrum of COVID-19: a population level analysis based on US National COVID Cohort Collaborative (N3C) data. The Lancet HIV, 8(11), e690-e700.

[68] Levitt, E. B., Patch, D. A., Mabry, S., Terrero, A., Jaeger, B., Haendel, M. A., ... & Johnson, J. P. (2022). Association Between COVID-19 and Mortality in Hip Fracture Surgery in the National COVID Cohort Collaborative (N3C): A Retrospective Cohort Study. JAAOS Global Research & Reviews, 6(1).

[69] Farhad, P., Greifer, N., Leyrat, C., Stuart, E. (2020). MatchThem:: matching and weighting after multiple imputation. arXiv:2009.11772 (2020). https://journal.r-project.org/archive/2021/RJ-2021-073/RJ-2021-073.pdf

[70] Clark, T. G., Altman, D.G. (2003). Developing a prognostic model in the presence of missing data: an ovarian cancer case study. J Clin Epidemiol., 56(1), 28–37.